\newcommand{\etal}{\textit{et al.}}
\newcommand{\eg}{\textit{e.g.}}
\newcommand{\ie}{\textit{i.e.}}
\newcommand{\vs}{\textit{vs.}}
\newcommand{\red}[1]{{\color{red}{#1}}} 
\newcommand{\blue}[1]{{\color{blue}{#1}}}
 
\newenvironment{packed_itemize}{
	\vspace{-0.15cm}\begin{itemize}
		\setlength{\itemsep}{1pt}
		\setlength{\parskip}{0pt}
		\setlength{\parsep}{0pt}
	}{\end{itemize}}

\documentclass[final]{cvpr}

\usepackage{times}
\usepackage{epsfig}
\usepackage{graphicx}
\usepackage{amsmath}
\usepackage{amssymb}

\usepackage{cuted}
\usepackage{capt-of}

\usepackage[pagebackref=true,breaklinks=true,colorlinks,bookmarks=false]{hyperref}

\def\cvprPaperID{3719} % *** Enter the CVPR Paper ID here
\def\confYear{CVPR 2021}

\begin{document}

%%%%%%%%% TITLE
\title{Removing Diffraction Image Artifacts in Under-Display Camera via \\Dynamic Skip Connection Network}

% \author{Ruicheng Feng\\
% Nanyang Technological University\\
% {\tt\small ruicheng002@e.ntu.edu.sg}
% \and
% Chongyi Li\\
% Nanyang Technological University\\
% {\tt\small chongyi.li@ntu.edu.sg}
% }

\author{
Ruicheng Feng$^{1}$\quad
Chongyi Li$^{1}$\quad
Huaijin Chen$^{2}$\quad
Shuai Li$^{2}$\quad
Chen Change Loy$^{1}$\quad
Jinwei Gu$^{2,3}$
\\
$^{1}$S-Lab, Nanyang Technological University\quad
$^{2}$Tetras.AI\quad
$^{3}$Shanghai Al Laboratory\\
{\tt\small \{ruicheng002, chongyi.li, ccloy\}@ntu.edu.sg}
\\
{\tt\small \{huaijin.chen, shuailizju\}@gmail.com\quad
gujinwei@tetras.ai}
}

\thispagestyle{empty}
\twocolumn[{
    \renewcommand\twocolumn[1][]{#1}
    \vspace{-1em}
    \maketitle
    \vspace{-0.9cm}
    \begin{center}
        \centering
        \includegraphics[width=0.98\textwidth]{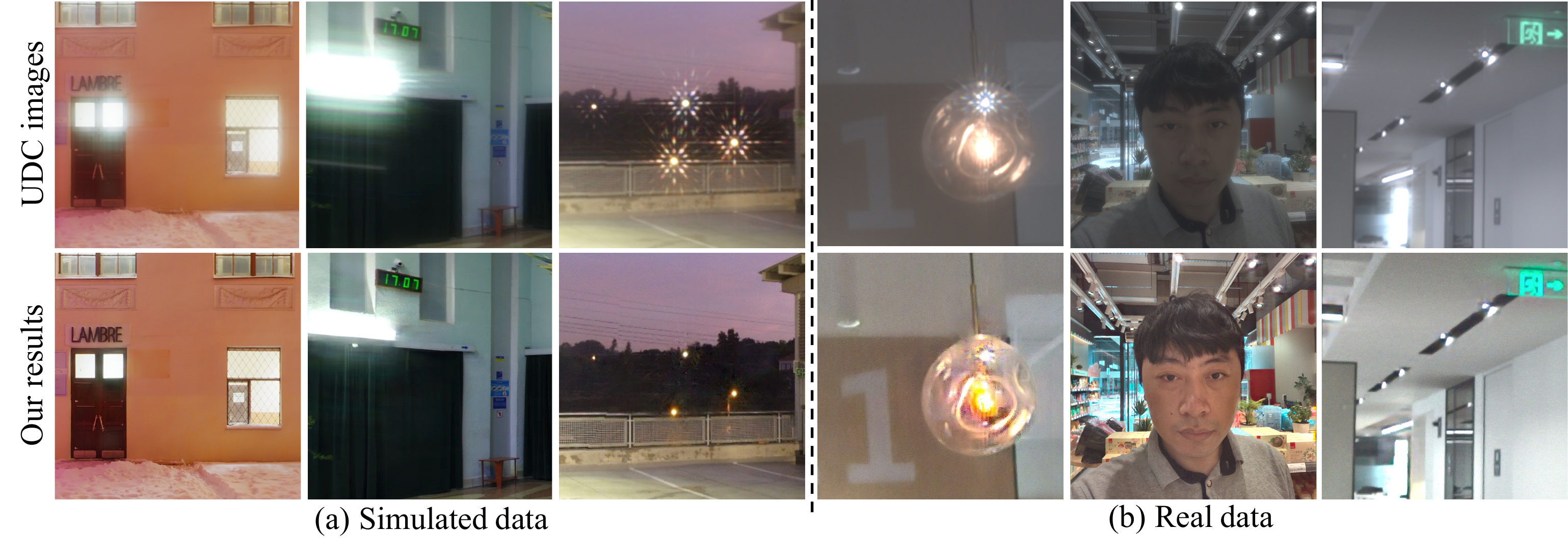}
        \vskip -0.2cm
        \captionof{figure}{\textbf{Removing Diffraction Artifacts from Under-Display Camera (UDC) images.} The major degradations caused by light diffraction, \eg, flare, blur, and haze, could significantly affect the visual quality of UDC images. Our method effectively restores fine details and suppresses the diffraction effects of UDC images.}
        \label{fig:fig_1}
    \end{center}
}]
% \thispagestyle{empty}
% \pagestyle{empty}

% \maketitle
% \begin{figure*}[!ht]
% \begin{center}
% % \fbox{\rule{0pt}{2in} \rule{0.9\linewidth}{0pt}}
%   \includegraphics[width=0.98\linewidth]{figs/fig_1.pdf}
% \end{center}
% \vskip -0.5cm
%   \caption{\textbf{Removing Diffraction Image Artifacts from UDC images.} Our method restores fine details and suppresses diffraction effects from both simulated and real data.}
% \label{fig:fig_1}
% \vskip -0.5cm
% \end{figure*}

% \vspace{-4mm}
%%%%%%%%% ABSTRACT
\begin{abstract}
\vskip -0.2cm
Recent development of Under-Display Camera (UDC) systems provides a true bezel-less and notch-free viewing experience on smartphones (and TV, laptops, tablets), while allowing images to be captured from the selfie camera embedded underneath. In a typical UDC system, the microstructure of the semi-transparent organic light-emitting diode (OLED) pixel array attenuates and diffracts the incident light on the camera, resulting in significant image quality degradation. Oftentimes, noise, flare, haze, and blur can be observed in UDC images. In this work, we aim to analyze and tackle the aforementioned degradation problems. We define a physics-based image formation model to better understand the degradation. In addition, we utilize one of the world's first commodity UDC smartphone prototypes to measure the real-world Point Spread Function (PSF) of the UDC system, and provide a model-based data synthesis pipeline to generate realistically degraded images. We specially design a new domain knowledge-enabled Dynamic Skip Connection Network (DISCNet) to restore the UDC images. We demonstrate the effectiveness of our method through extensive experiments on both synthetic and real UDC data. Our physics-based image formation model and proposed DISCNet can provide foundations for further exploration in UDC image restoration, and even for general diffraction artifact removal in a broader sense.
\footnote{Codes and data are available at \href{https://jnjaby.github.io/projects/UDC}{https://jnjaby.github.io/projects/UDC}.}
% Codes and data are available at \href{https://jnjaby.github.io/projects/UDC}{https://jnjaby.github.io/projects/UDC}.
\end{abstract}

%%%%%%%%% BODY TEXT
\vspace{-0.2cm}
\section{Introduction}

\begin{figure}[t]
\centering
  \includegraphics[width=.96\linewidth]{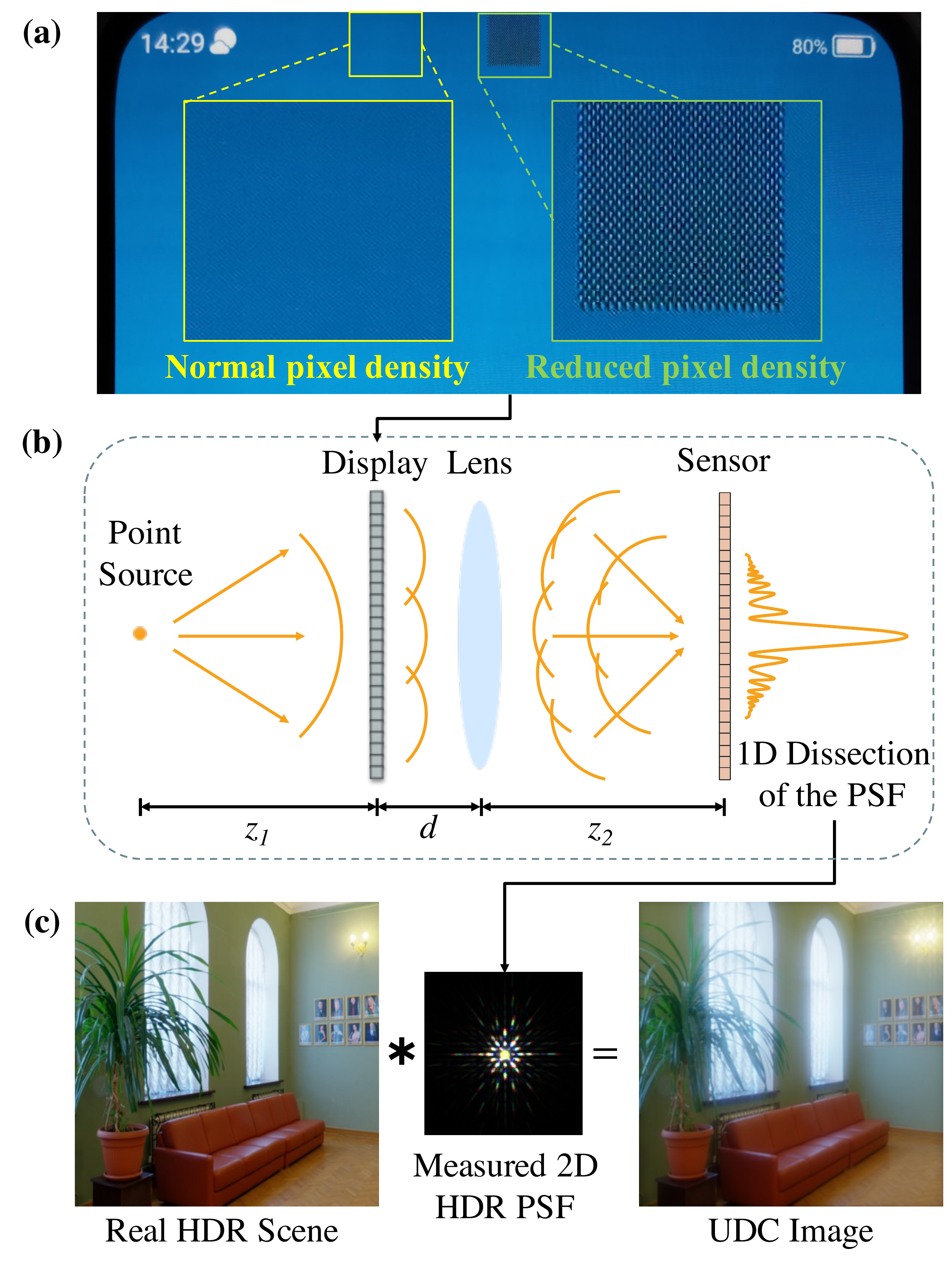}
  \caption{\textbf{(a)} A close-up shot of the UDC OLED on the ZTE Axon 20 phone. The UDC OLED panel has reduced pixel density in the area above the camera, allowing for more transparency. \textbf{(b)} The schematic of the UDC system. The light emitted from a point light source is modulated by the OLED and camera lens, before being captured by the sensor. \textbf{(c)} A simulated example of our image formation model with a real-captured PSF.  The 2D PSF is brightened in the figure to visualize the structured sidelobe patterns.}
  \vspace{-0.3cm}
\label{fig:system}
\end{figure}

The consumer demand for smartphones with bezel-free, notch-less display has sparked a surge of interest from the phone manufacturers in a newly-defined imaging system, Under-Display Camera (UDC).
Besides smartphones, UDC also demonstrates its practical applicability in other scenarios, \ie, for videoconferencing with UDC TV, laptops, or tablets, enabling more natural gaze focus as they place cameras at the center of the displays \cite{msudc2020}. 
As Figure \ref{fig:system} shows, a typical UDC system has the camera module placed underneath and closely attached to the semi-transparent Organic Light-Emitting Diode (OLED) display.  
% In particular, embedding a camera under the display enables mobile manufacturers to devise phones with a true full-screen design without any notches or cutouts. The imaging formation is schematically illustrated in Figure \ref{fig:formation}.
Although the display looks partially transparent, the regions where the light can pass through, \ie the gaps between the display pixels, are usually in the micrometer scale, which substantially diffracts the incoming light \cite{qin2016see}, affecting the light propagation from the scene to the sensor. 
The UDC systems introduce a new class of complex image degradation problems, combining strong flare, haze, blur, and noise (see top row in Figure \ref{fig:fig_1}). In the first attempt \cite{zhou2020image} to address the UDC image restoration problem, the authors proposed a Monitor-Camera Imaging System (MCIS) to capture paired data and used the image formation to synthesize the Point Spread Function (PSF) of two types of OLED display. However, there are several drawbacks of this pioneer work, including 1) inaccurate PSF due to a mismatch between the actual and the synthetic PSF, 2) lacking proper high dynamic range (HDR) in the MCIS-captured data, missing realistic UDC degradation, 3) prototype UDC differs significantly from actual production UDC, 4) missing real-world evaluation on non-MCIS data, and 5) proposed network does not fully utilize domain knowledge. We examine the drawbacks in further details in Section \ref{sec:relate}. 
% : it failed to consider one of the most prominent artifacts, diffraction flare, due to the limited dynamic range of the MCIS system, 2) the OLED display, and inherently, the PSF used in their experiments, is far from the actual production UDC OLED, and 3) the UDC setup in their experiment is not rigidly assembled together; A nuanced change in the position between the display and the camera (i.e. slight rotation or displacement) will lead to significantly different diffraction PSF. As a result, the artifacts on their UDC images are significantly different than actual UDC images. Thus, the model trained in \cite{zhou2020image} does not work well on actual images captured on real-world UDC devices.

In this work, we aim to address the aforementioned issues. We first present a realistic image formation model and measurement protocol considering proper dynamic range for the scenes and camera sensor, and restore the real-world degradation in the actual UDC images.
% using our novel domain knowledge-enabled DISCNet. 
To this end, we experiment with one of the world's first production UDC device, ZTE Axon 20, which incorporates a UDC system into its selfie camera. 
% Such OLED has reduced pixel density in the area that covers the camera (see Figure \ref{fig:udc_closeup}), allowing higher light transmittance than regular OLED panels and changes the diffraction PSF shape significantly from a regular OLED.
Note that we aim to analyze and investigate the artifacts caused by diffraction effects, rather than propose a product-ready solution for ZTE phone camera.
Our method is versatile and applicable to other UDC device, or more generally, other diffraction-limited imaging systems, \eg, microscopy imaging, pinhole camera.
% Note that we use ZTE Axon 20 because it is the only publicly available UDC system. Our method is versatile and is applicable to other UDC device.
%
We devise an imaging system to directly measure the PSF of the UDC device (see Section \ref{sec:psf}) with a point source. As shown in Figure \ref{fig:system}, due to the diffraction of the display, the resulting PSF has some special characteristics: it has large spatial support, strong response at the center, and long-tail low-energy sidelobes.
With the measured PSF, we reformulate the image formation model to account for realistic flare, haze, and blur, which were missing \cite{zhou2020udc, zhou2020image} due to the limited dynamic range of scenes. 
% Unlike \cite{zhou2020image}, we propose to use real-world-level irradiance with high dynamic range (HDR) to model the diffraction flare effects and introduce sensor saturation into the imaging system. 
Then, we develop a data simulation pipeline based on the image formation model by using HDR images to approximate real scenes.
Additionally, we capture real images using the UDC phone's selfie camera to validate our simulated data and evaluate the performance of our restoration network. As shown in Figure \ref{fig:fig_1}, our simulated and real data reveal similar degradation, especially in those high-intensity regions. Specifically, \textit{flare} can be observed nearby strong light sources, where highlights are spread into neighboring low-intensity areas in structured diffraction patterns. 
% Note that the image formation model cannot completely reflect the real-world imaging process, as it involves more complicated degradations. Still, data simulation based on this model can enable us better understand and quantitatively analyze the major degradation.
% Although recent advances in deep restoration models have boost performance on image restoration applications, such as denoising \cite{zhang2017learning,zhang2018ffdnet,guo2019toward}, debluring \cite{kupyn2018deblurgan,kupyn2019deblurgan}, superresolution \cite{dong2014learning,wang2018esrgan,zhang2019ranksrgan}, and low-light enhancement \cite{wei2018deep,gharbi2017deep,chen2018learning,guo2020zero}, simply applying existing end-to-end methods on these low-quality images with complicated degradation types can only achieve performance far below satisfaction. 
% Instead, we consider a non-blind setting and tailor the network architecture to the recovery from such severely degraded patterns. 

To restore the UDC images, we propose a DynamIc Skip Connection Network (DISCNet) that incorporates the domain knowledge of the image formation model into the network designs. In particular, sensor saturation breaks the shift-invariance of the single-PSF-based convolution, leading to spatially-variant degradation. This motivates us to design a dynamic filter network to dynamically predict filters for each pixel. In addition, due to large support of PSF, we propose a multi-scale architecture and perform dynamic convolution in the feature domain to obtain a larger receptive field. Also, a condition encoder is introduced to utilize the information of PSF.  

In summary, our contributions are as follows:
\begin{packed_itemize}
    \item We reformulate the image formation model for UDC systems by considering dynamic range and saturation, which takes into account the diffraction flare commonly seen in UDC images.
    \item We utilize the first UDC smartphone prototypes to measure the real-world PSF. The PSF is used as part of a model-based data synthesis pipeline to generate realistic degraded images.
    \item We devise a DynamIc Skip Connection Network (DISCNet) that incorporates the domain knowledge of the UDC image formation model. Experimental results show that it is effective for removing diffraction image artifacts in UDC systems.
\end{packed_itemize}

%-------------------------------------------------------------------------
%----------- Related Work ------------------------------------------------
%-------------------------------------------------------------------------
\section{Related Work}
\label{sec:relate}
\noindent\textbf{UDC Imaging.}
Several previous work \cite{qin2017evaluation, tang202028} characterized and analyzed the diffraction effects of UDC systems. Kwon \etal \cite{kwon2016modeling} modeled the edge spread function of transparent OLED. Qin \etal \cite{qin2016see} discussed pixel structure design that can potentially reduce the diffraction. While all these works provide good insights into UDC imaging systems, none of them tackles the image restoration problem. Additionally, several works \cite{hirsch2009bidi,suh2012p, suh201350} proposed camera-behind-display design for enhanced 3D interaction with flat panel display. Though low-resolution images are the by-products of those prototype interaction systems, given the extremely poor image quality, they are unsuitable for daily photography, which is the focus of this work.

\noindent\textbf{UDC Restoration.}
To our best knowledge,  \cite{zhou2020image} and the subsequent ECCV challenge \cite{zhou2020udc} are the only works that directly address the problem of UDC image restoration. In \cite{zhou2020image}, the authors devised an MCIS to capture paired images, and solve the UDC image restoration problem as a blind deconvolution problem using a variant of UNet \cite{ronneberger2015u}. While the work pioneers the UDC image restoration problem, it suffers from several drawbacks.

\textit{First}, while MCISs are commonly used in the computational imaging community \cite{peng2019learned, asif2016flatcam} to capture the system PSF or acquire paired image data, most commodity monitor lacks the high dynamic range which is a must to model realistic diffraction artifacts in UDC systems. As a result, the PSFs they used have incomplete side lobes, and the images have less severe artifacts, \eg, blur, haze, and flare. In our work, we consider HDR in data generation and PSF measurement to allow us to tackle real-world scenes properly. 
\textit{Secondly}, the authors use regular OLED manually covering a camera in their setup, instead of an actual rigid UDC assembly, and perform experiments and evaluations on quasi-realistic data. 
As a result, any slight movements, rotation, or tilt of the display with respect to the sensor plane will cause variational PSFs, preventing their network from being applied to handle variational degradations without the knowledge of the PSF kernel. 
% the authors used two regular general-purpose OLED display panels to generate the datasets, which have different pixel structure (consequently vastly different PSF) and significantly lower light transmittance 
% ($20\%$ for T-OLED and $2.9\%$ for P-OLED, which are very different than the ~30\% in production UDC OLED panels) 
% Additionally, the authors manually cover the camera with the display for the data collection. At test time, any small movements, rotation or tilt of the display with respect to the camera will cause very different diffraction PSF. 
% Also, they perform experiments and evaluations on quasi-realistic data, where the target scenes are images displayed on flat-panel display. 
To minimize the domain gap, we use one of the world's first production UDC device for data collection, experiments, and evaluations.
\textit{Lastly}, though the authors captured and used the PSF in data synthesis, they formulated the UDC image restoration as a blind deconvolution problem through a simple UNet, without explicitly utilizing the PSFs as useful domain knowledge. In contrast, we leverage the PSF as important supporting information in our proposed DISCNet.
% In other word, despite having the PSF, 
% Furthermore, the present data synthesis pipeline takes a low dynamic range (LDR) image and an approximated blur kernel as inputs, which fails to take into account the diffraction effects and blur degradations.

\noindent\textbf{Non-blind Image Restoration.} In the context of non-blind image restoration, a large body of works has exerted great effort to tackle this ill-posed problem. Prior to the deep-learning era, early deconvolution approaches \cite{richardson1972bayesian,orieux2010bayesian,levin2009understanding,cho2011handling,whyte2014deblurring} imposed prior knowledge to constrain the solution space since the noise model is unknown. Then, several works \cite{schuler2013machine,xu2014deep,zhang2017learning} focused on establishing the connection between optimization-based deconvolution and a neural network for non-blind image restoration.
Also, Shocher \etal \cite{shocher2018zero} employed a small image-specific network to deal with various degradations of a single image. Zhang \etal \cite{zhang2018learning} proposed SRMD to handle multiple degradations with one network. Gu \etal \cite{gu2019blind} proposed SFTMD and Iterative Kernel Correction (IKC) to iteratively correct the kernel code of degradations. Additionally, \cite{bell2019blind,yuan2018unsupervised,zhou2019kernel} used Generative Adversarial Networks (GANs) to tackle different degradations. Similar to SRMD \cite{zhang2018learning}, we take the PSF kernel as an additional condition but use it in a different way, \ie, feed it into a condition encoder to facilitate dynamic filter generation.

\noindent\textbf{Dynamic Filter Network.} Recent years have witnessed great success in dynamic filter networks employed in a wide range of vision applications to handle spatially-variant issues. Jia \etal \cite{jia2016dynamic} firstly exploited dynamic network to generate an individual kernel for each pixel conditioned on the input image. Since then, this module has proven to provide significant benefits for applications, such as video interpolation \cite{niklaus2017video, niklaus2017video2}, denoising \cite{bako2017kernel,mildenhall2018burst,xu2019learning}, super-resolution \cite{jo2018deep,xu2020unified,wang2018recovering}, and video deblurring \cite{zhou2019spatio}. In addition, Wang \etal \cite{wang2019carafe} proposed a kernel prediction module serving as a universal upsampling operator. Most previous approaches, however, can not be directly applied to UDC image restoration, because they either apply predicted filters in the image domain or mainly focus on a special operation. In this work, we construct multi-scale filter generators and adopt the dynamic convolution in the feature domain to handle degradation with large-support and long-tail PSF.

%-------------------------------------------------------------------------
%----------- Image Formation and Dataset ---------------------------------
%-------------------------------------------------------------------------
\vspace{-1mm}
\section{Image Formation Model and Dataset}
\label{sec:imaging}
\vspace{-1mm}
\subsection{Image Formation Model}
\vspace{-1.5mm}
We consider a real-world image formation model for UDC that suffers from several types of degradation, including diffraction effects, saturation, and camera noise. This degradation model is given by
\begin{equation}
\label{eqn:formation}
    \hat{y} = \phi[C(x*k+n)],
\end{equation}
where $x$ represents the real scene irradiance that has a high dynamic range (HDR). $k$ is the known convolution kernel, commonly referred to as the Point Spread Function (PSF), $*$ denotes the 2D convolution operator, and $n$ models the camera noise. To model saturation derived from the limited dynamic range of digital sensor, we apply a clipping operation $C(\cdot)$, formulated by $C(x)=\text{min}(x, x_{max})$, where $x_{max}$ is a range threshold. A non-linear tone mapping function $\phi(\cdot)$ is used to match the human perception of the scene.
% Next, we measure and analyze the optical properties of the PSF.

% \begin{figure}[t]
% \begin{center}
%   \includegraphics[width=0.98\linewidth]{figs/hdr_syn.pdf}
% \end{center}
%     \vskip -0.2cm
%   \caption{Comparison on images simulated with LDR and HDR.}
% \label{fig:hdr_syn}
% \end{figure}

\begin{figure}[t]
\centering
  \includegraphics[width=0.98\linewidth]{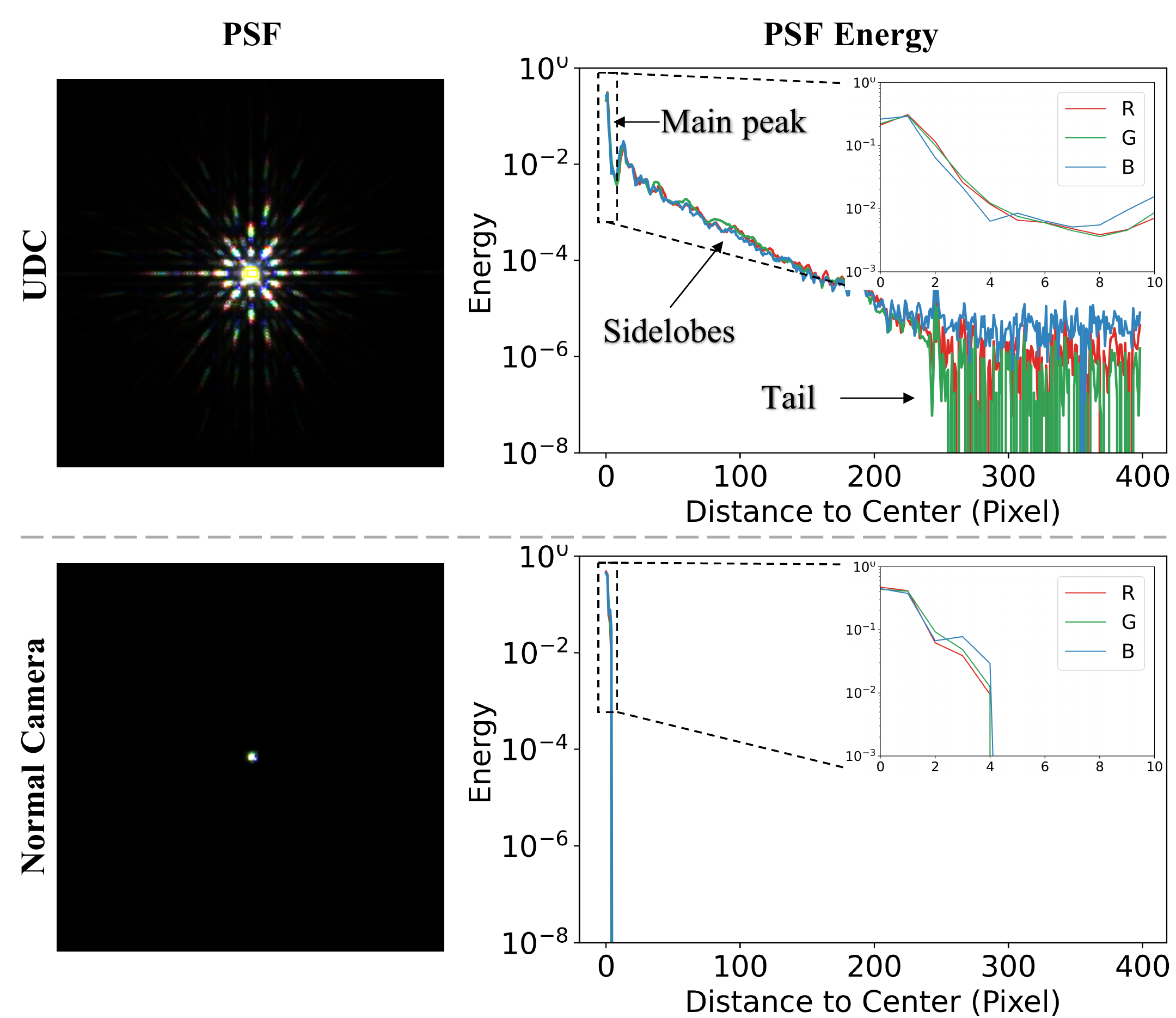}
\caption{\textbf{Comparison of PSF energy for UDC and normal camera.} The PSF is brightened to visualize the structured sidelobe patterns. Due to the finite aperture size and manufacturing imperfections, the PSF of a real normal camera (Bottom left) will be a blur kernel of some size, rather than a perfect point.}
\label{fig:psf_energy}
\vskip -0.5cm
\end{figure}
% \vspace{-2mm}
%-------------------------------------------------------------------------
\subsection{PSF Measurement} 
\label{sec:psf}
In Figure \ref{fig:system}, the optical field $U_{S}(p,q)$ captured by the sensor given a unit amplitude point source input can be expressed as
\begin{equation}
\label{eqn:prop}
\begin{aligned}
    U_{S}(p,q)=&\left\{\left[\exp\left({\frac{i\pi r^{2}}{\lambda z_{1}}}\right)\cdot t(p,q)\right]*\exp\left({\frac{i\pi r^{2}}{\lambda d}}\right)\right.\\
    &\left.\cdot\exp\left({\frac{-i\pi r^{2}}{\lambda f}}\right)\right\}*\exp\left({\frac{i\pi r^{2}}{\lambda z_{2}}}\right).
\end{aligned}
\end{equation}
Here, $(p,q)$ is the 2D spatial coordinates, $r^{2}=p^{2}+q^{2}$, $\lambda$ is the wavelength, $f$ is the focal length of the lens, $t(p,q)$ is the transmission function of the display. $z_{1}$, $d$ and $z_{2}$ denote the distance between the light source and the display, the distance between the display and the lens, and the distance between the lens and the sensor, respectively. $*$ denotes the convolution, and $\cdot$ denotes multiplication. Finally, the PSF of the imaging system is given by $k=|U_{S}|^{2}$.

% However, as shown in the Supplemental Material, although the simulated PSF has a very similar shape as the real PSF, differences still exist due to model approximations and manufacturing imperfections. In addition, due to proprietary reasons, we do not have access to the detailed transmission function $t(p,q)$ for the UDC device. 

With the exact pixel layout of a certain display, we can theoretically simulate the PSF of an optical system modulated by the display. However, we found that although the simulated and real-measured PSF share a similar shape, they slightly differ in color and contrast due to model approximations and manufacturing imperfections (see Supplement for light propagation model and simulated PSF). Besides, we have no access to the transmission function $t(p,q)$ for the OLED display we used in this work, whose pixel structure is unknown due to proprietary reasons.

Therefore, we follow \cite{Sun_2020_LearnedOpticHDR} and devise an imaging system to directly measure the PSF by placing a white point light source $1$-meter away from the OLED display. At this distance, the size of the point light source is equivalent to one pixel of the sensor.
Hence, this illuminant can be considered as an impulse input.
To capture the entire PSF, including the strong main peak and the weak sidelobes, we take three images successively at different exposures: [$1$, $1/32$, $1/768$], 
which are then normalized to the same brightness level. Subsequently we pick out all unsaturated pixel values to fuse into one HDR image.
The captured PSF of the UDC system (Figure \ref{fig:psf_energy} top) shows structured patterns: 1) the response at the center, denoted as main peak, is very strong and has greater energy with an order of magnitude. 2) Compared to the PSF of a normal camera, it has larger spatial support (over $800\times800$) and spike-shaped long-tail sidelobes whose energy decreases exponentially. 3) In the tail regions of the sidelobe, we can an observe obvious color shift. To summarize, the PSF of UDC has several special characteristics compared to regular blur kernel, which motivates a simulation based on HDR images.

Compared with the UDC image formation model described in \cite{zhou2020image}, our model is closer to the real situation in the following two aspects. First, the objects $x$ that we considered are real scenes with high dynamic range.
Since the PSF of UDC has a strong response at the center but vastly lower energy at long-tail sidelobes, only when convolved with sufficiently high-intensity scenes, these spike-shaped sidelobes can be amplified to be visible (flares) in the degraded image. Hence, images captured by UDC systems in real scenes will exhibit structured flares near strong light sources. The imaging system in \cite{zhou2020image}, however, cannot model this degradation, because it captures images displayed on an LCD monitor, which commonly has limited dynamic range.
We demonstrate in Supplement that if we clip the same scene from high dynamic range to low dynamic range, these flares caused by diffraction become invisible. 
Second, due to the high dynamic range of the input scene, the digital sensor (usually 10-bit) will inevitably get saturated in real applications, resulting in an information loss. This factor should be considered in the image formation model as well.

%-------------------------------------------------------------------------
% \vspace{-2mm}
\subsection{Data Collection and Simulation}
\label{sec:data}
% \vspace{-2mm}
\noindent\textbf{Simulated Data.} To generate the synthetic data, we gathered $132$ HDR images with large dynamic ranges from HDRI Haven dataset\footnote{\href{https://hdrihaven.com/hdris/}{https://hdrihaven.com/hdris/}.}. Each HDR panorama image is a 360-degree panorama of resolution $8192\times4096$. We first re-projected these panorama images back to perspective view and then cropped them into $800\times800$ patches. In this way, we got a total of $2016$ subimages for training and $360$ for testing. For each of the crops, we simulated the corresponding degraded image using Eqn. \ref{eqn:formation}, where the PSF calibrated in Section \ref{sec:psf} is used as the kernel $k$. Refer to the Supplemental Material for more details.

\noindent\textbf{Real Data.} For each real scene, we captured three images of different exposures: [$1$, $1/4$, $1/16$] using ZTE Axon 20 phone, and then combine them into one HDR image. To ensure the linearity of the data, we directly used the raw data after HDR fusion, without any non-linear processing.

%-------------------------------------------------------------------------
%----------- Methodology -------------------------------------------------
%-------------------------------------------------------------------------
% \vspace{-1mm}
\section{Dynamic Skip Connection Network}
% \vspace{-2mm}
%-------------------------------------------------------------------------
\subsection{Motivation}
\label{sec:motivation}
% \vspace{-1mm}
We treat UDC image restoration as a non-blind image restoration problem, where a degraded image $\{\hat{y}_i\}$ and the ground-truth degradation (PSF) $\{k_i\}$ are given to restore the clear image $\{x_i\}$. In general, with the known convolution kernel, non-blind restoration establishes the upper bound for blind restoration, where the kernel needs to be estimated. Despite claiming our method as non-blind, we note that it can be used towards blind UDC image restoration by incorporating any PSF estimation algorithm. 
% Non-blind image restoration provides an upper bounds for blind restoration where the degradation are unknown and required estimation. Note that any estimation approaches are orthogonal to our method and can be easily extended on blind setting. 

Traditionally, non-blind image restoration is solved by classical deconvolution, \eg, Wiener filter \cite{orieux2010bayesian}, which have a rigorous assumption on the linearity of the system. UDC artifacts occur in HDR scenes, where the sensor is over-saturated in high-intensity area, breaking the linearity of the system and losing the information within.
Additionally, traditional deconvolution do not consider extremely large kernels ($800\times800$),
thus causing serious ringing and halo artifacts (Figure \ref{fig:syn_visual} and Figure \ref{fig:real_visual}).
Moreover, deep learning-based methods could leverage more data to learn restoration and require only one forward pass during inference.
% which may not hold for high dynamic range scenarios where the UDC artifacts occur. Sensor saturation not only breaks the linearity of the system and loses information $\hat{y}_i$ and its corresponding kernel $k_i$, it is intractable to restore $x_i$, since the information in saturated regions is irrevocably lost
% in the linear domain (without tone mapping). However, these approaches suffer from serious ringing and halo artifacts, as shown in Figure \ref{fig:syn_visual} and Figure \ref{fig:real_visual}. This is because: 1) optimization in linear domain is typically dominated by the errors in high-intensity regions, where pixels have values with orders of magnitude greater than those in darker regions when measured in the linear domain, and 2) even with observation $\hat{y}_i$ and its corresponding kernel $k_i$, it is intractable to restore $x_i$, since the information in saturated regions is irrevocably lost.
In this regard, we use a network to reconstruct $\hat{x}_i=\phi(x_i)$, which suggests a recovery from $\hat{y}$ to $\hat{x}$ in the non-linear tone-mapped domain, yielding triplet set $\{\hat{y}_i, k_i, \hat{x}_i\}$. Such optimization in the tone-mapped domain gives more emphasis to darker pixels and encourages the balance of restoration in different regions.

% \vspace{-0.5mm}
Moreover, the image formation model in Eqn. \ref{eqn:formation} assumes a shift-invariant 2-D convolution. Now in the tone-mapped domain with non-linear sensor saturation, such assumption no longer holds, since the PSF's shape and intensity can be variant based on the input pixel and its neighborhood at the corresponding location. For example, the OLED diffracting saturated highlights into neighboring unsaturated areas motivates an adaptive recovery of clipped information from the nearby areas.
% is variant depending on the depth of the scene and spatial location on the sensor.
% The image formation model in Eqn. \ref{eqn:formation} assumes a shift-invariant 2-D convolution, but this property does not hold for the full imaging process due to sensor saturation. Hence, the real degradations can be complicated and varies spatially, especially in those highlight regions. In addition, the optical feature of an OLED display that spreads out saturated highlights into neighboring unsaturated areas motivates an adaptive recovery of clipped information from the nearby areas.
Inspired by recent success of Kernel Prediction Network (KPN) \cite{jia2016dynamic, mildenhall2018burst, niklaus2017video, zhou2019spatio}, we propose DynamIc Skip Connection Network (DISCNet), which dynamically generates filter kernel at each pixel and applies them to different feature spaces at different network layers with skip connections. This network is conditioned on two inputs: 1) the PSF that provides domain knowledge about the image formation model, and 2) the degraded image that provides light intensity and neighborhood context information to facilitate a spatially-variant recovery. We demonstrate the effectiveness of the coupled conditions in Section \ref{sec:ablation}.

For dynamic convolution, directly applying the predicted filters in the image domain like most existing KPN-based approaches is not best suited for UDC image restoration, because the PSF in UDC has large support and long-tail side lobes (see Figure \ref{fig:psf_energy}). As discussed in \cite{xu2014deep}, such an inverse convolution process with a large PSF can only be well approximated in image domain with sufficiently large kernels  (larger than $100$), while the size of dynamic filters is typically far smaller (\eg $5$ or $7$). Therefore, we propose to apply dynamic convolution in the feature domain. On top of that, we construct a multi-scale architecture, where the filter generator at each scale predicts dynamic filters separately, to further enlarge the spatial support of the learned filters.

% To further boost the performance of filter adaptive dynamic convolution, an intuitive way is to enlarge the filter size. However, the computation complexities grows quadratically when increasing filter size $s$, leading to high memory cost and strongly limiting the applicability in practical scenarios.

%-------------------------------------------------------------------------
\subsection{Network Architecture}
\begin{figure*}[t]
\centering
  \includegraphics[width=0.85\linewidth]{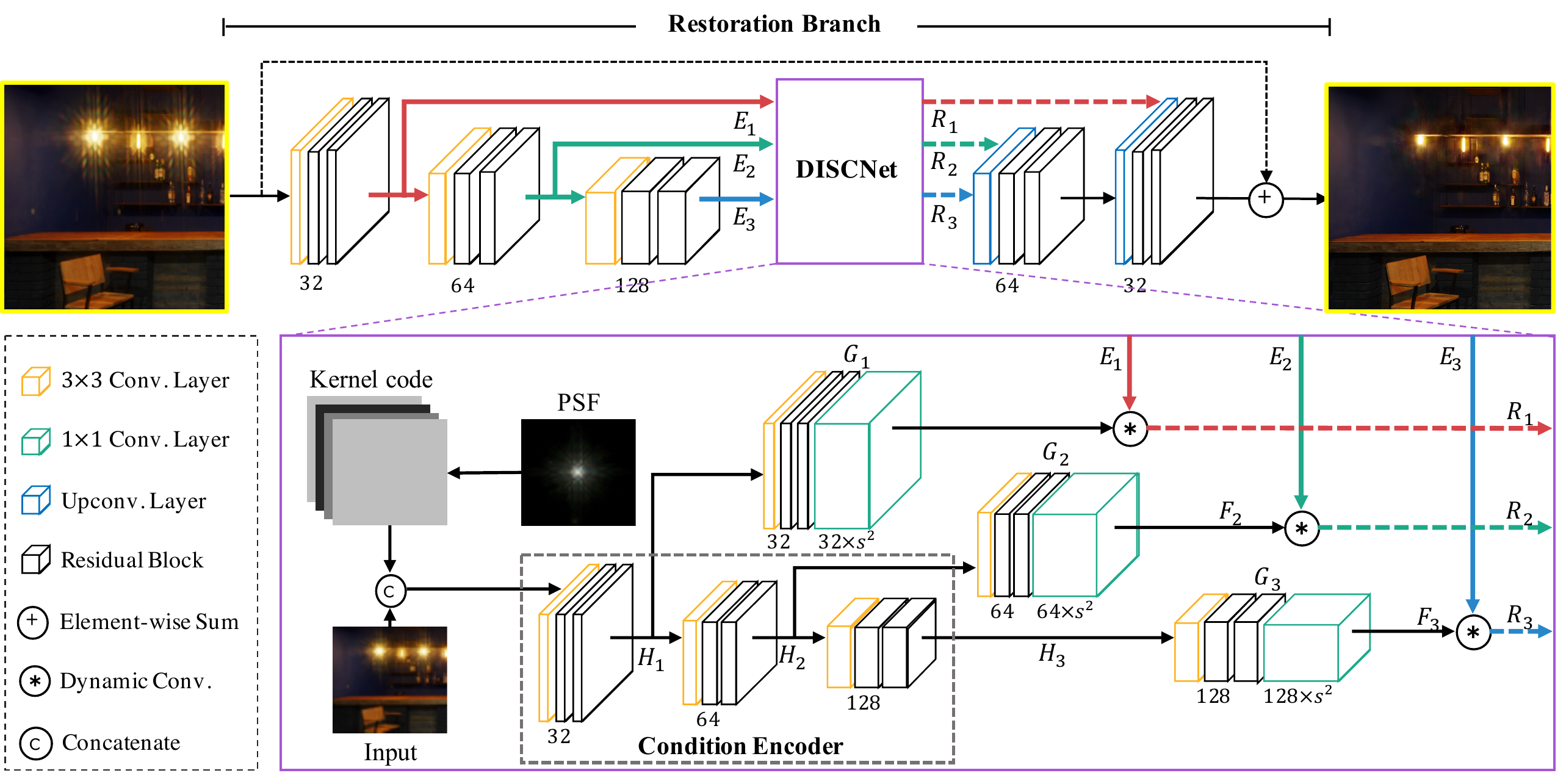}
\caption{\textbf{Illustration of the proposed DISCNet.} The main restoration branch consists of an encoder and a decoder, with feature maps propagated and transformed by DISCNet through skip connections. DISCNet applies multi-scale dynamic convolutions using generated filters conditioned on PSF kernel code and spatial information from input images.}
\label{fig:structure}
\vskip -0.3cm
\end{figure*}

As shown in Figure \ref{fig:structure}, our network comprises a restoration branch and a DynamIc Skip Connection Network (DISCNet). The restoration branch learns to extract features and restore the final clean image. DISCNet is employed to tackle various degradations and transform and refine the features extracted from the restoration branch.

\noindent\textbf{Training with Various Degradations.} Suppose the degraded image $\hat{y}_i$ is of shape $H\times W\times C$, where $H$, $W$, $C$ denote the height, width, and the number of channels of images. Following \cite{gu2019blind}, we project the PSF onto a $b$-dimensional vector, referred to as kernel code, by Principal Component Analysis (PCA) to reduce computational complexities. The kernel code is then stretched into degradation maps of size $H\times W\times b$ and concatenated with the degraded image to get the condition maps of size $H\times W\times (b+C)$ , which are then fed into the DISCNet. In this paper, we empirically set $b=5$.

\noindent\textbf{Restoration Branch.} This branch builds upon an encoder-decoder architecture with skip connections to restore the degraded images. Specifically, the encoder contains three convolutional blocks, each of which has a $3\times3$ convolution layer with stride $2$, a LeakyReLU \cite{he2015delving} layer, and two residual blocks \cite{he2016deep}, extracting features $E_1$, $E_2$, $E_3$ at three different scales. The extracted features are fed into DISCNet and transformed into $R_1$, $R_2$, $R_3$, respectively. Similarly, the decoder consists of two convolutional blocks, including an up-convolution layer and two residual blocks. Each convolutional block takes the transformed feature at its corresponding scale as input and reconstructs the final tone-mapped sharp images.

\noindent\textbf{DynamIc Skip Connection Network.} The proposed DISCNet mainly consists of three designs: condition encoder,  multi-scale filter generator, dynamic convolution.

Given the condition maps as input, the condition encoder extracts scale-specific feature maps $H_1$, $H_2$, $H_3$ using $3$ blocks similar to the encoder of the restoration branch. 
Although the kernel code maps are globally uniform, the condition encoder could still capture rich information from the degraded image with spatial variability and manage to recover saturated information from nearby low-light regions.

Then, the extracted features at different scales are fed into their corresponding filter generators, where each comprises a $3\times3$ convolution layer, two residual blocks, and a $1\times1$ convolution layer to expand feature dimension. Particularly, given the size of dynamic filters $s$, a filter generator $G_n$ takes in the extracted feature maps at a specific scale $H_n\in\mathbb{R}^{h\times w\times c}$ and outputs predicted filters $F_n=G_n(H_n)$, where the generated filters $F_n$ is of size $h\times w\times cs^2$. The filters are then used by a dynamic convolution to refine features $E_n$. For each pixel $(i,j,c_m)$ of features $E_n\in h\times w\times c$, the output feature $R_n$ is given by
\begin{equation}
    R_n(i,j,c_m)=\langle K_n(i,j,c_m),\varphi(E_n(i,j,c_m))\rangle,
\end{equation}
where $K_n(i,j,c_m)$ is a $s\times s$ filter reshaped from $F_n(i,j,c_m)\in\mathbb{R}^{1\times 1\times s^2}$. $\varphi(\cdot)$ denotes a $s\times s$ patch centered at position $(i,j,c_m)$, and $\langle\cdot\rangle$ represents inner product. The refined feature $R_n$ is then cast to the restoration branch.

%-------------------------------------------------------------------------
%----------- Experiments -------------------------------------------------
%-------------------------------------------------------------------------
\vspace{-0.1cm}
\section{Experiments}
\begin{table*}[!ht]
  \centering
  \caption{\textbf{Ablation results on the simulated dataset.} Starting from the baseline model, we gradually add each component in our network to validate their effectiveness. ``*'' indicates results evaluated on the simulation over single PSF. The best results are \textbf{highlight}.}
  \label{tab:ablation}
  \scalebox{0.96}{
  \begin{tabular}{l|ccc|cc|cc}
  	\hline\hline
  	Method & PSF & Filter Generators & Conditions & PSNR* & $\text{PSNR}_{avg}$ & SSIM* & $\text{SSIM}_{avg}$\\\hline
  	(a) Baseline on $1$ kernel & Single & - & - & $41.47$ & $38.55$ & $0.9850$ & $0.9742$ \\
  	(b) Baseline on various kernels & Variational & - & - & $40.67$ & $40.87$ & $0.9823$ & $0.9833$\\
    (c) w/ image conditions & Variational & Single-scale & Image & $41.33$ & $41.59$ & $0.9842$ & $0.9851$\\
    (d) w/ PSF conditions & Variational & Single-scale & PSF & $41.95$ & $42.14$ & $0.9848$ & $0.9857$\\
    (e) w/ image \& PSF conditions & Variational & Single-scale & Image + PSF & $42.60$ & $42.77$ & $0.9861$ & $0.9870$ \\
    (f) DISCNet (Ours) & Variational & Multi-scale & Image + PSF & $\mathbf{43.06}$ & $\mathbf{43.27}$ & $\mathbf{0.9870}$ & $\mathbf{0.9877}$ \\
  	\hline\hline
  \end{tabular}}
  \vskip -0.3cm
\end{table*}

%------------------------------------------------------------------------
\subsection{Implementation Details}
\noindent\textbf{Datasets.} We train the proposed model with the synthetic triplet data. To evaluate the effectiveness of DISCNet for non-blind degradations, we consider rotating PSF, which is analogous to rotating the display around the optical axis in imaging systems. To account for variations in the rotation angle, we build a kernel set in which the angles vary within $(-12, 12)$ where $0$ radian refers to the original PSF. Under this setting, each degraded image $\hat{y}_i$ is simulated using Eqn. \ref{eqn:formation}, with the convolution kernel $k_i$ is uniformly sampled from the kernel set. During training, the subimages are randomly cropped into $256\times256$ patches. More details about simulation settings can be found in Supplement Material.

\noindent\textbf{Training Setups.} We initialize all networks with Kaiming Normal \cite{he2015delving} and train them using Adam optimizer \cite{kingma2014adam} with $\beta_1=0.9$, $\beta_2=0.999$ and $\theta=10^{-8}$ to minimize a weighted combination of $\mathcal{L}_1$ loss and VGG loss \cite{johnson2016perceptual}. The mini-batch size for all the experiments is set to 16.
%The learning rate is initialized as $2\times10^{-4}$ and then decayed by half every $2\times10^5$ iterations.
The learning rate is decayed with a cosine annealing schedule, where $\eta_{min}=1\times10^{-7}$, $\eta_{max}=2\times10^{-4}$, and is restarted every $2\times10^5$ iterations.
For all experiments, we implement our models with the PyTorch \cite{paszke2017automatic} framework and train them using 2 NVIDIA V100 GPUs.

%------------------------------------------------------------------------
\subsection{Ablation Study}
\label{sec:ablation}
In this subsection, we analyze the effectiveness of each component in DISCNet. The baseline methods (Table \ref{tab:ablation}(a) and (b)) strip DISCNet in Figure \ref{fig:structure}. In this case, the restoration branch reduces to a variant of UNet architecture \cite{ronneberger2015u}, and $E_1$, $E_2$, $E_3$ are equivalent to $R_1$, $R_2$, $R_3$, respectively. Then we gradually apply different filter generators and condition maps for ablation studies. We report PSNR, SSIM, and LPIPS \cite{zhang2018unreasonable} as the evaluation metrics. The FLOPs is calculated by input size of $800\times800\times3$.

\noindent\textbf{Learning Variational Degradations.} Comparing Table \ref{tab:ablation}(a) and Table \ref{tab:ablation}(b), we found that our baseline trained on a dataset with only $1$ kernel can easily overfit to single degraded dataset but fails to generalize to other degradation types. In particular, the performance deteriorates seriously across other datasets, due to the discrepancy between the assumed PSF and real ones.

\noindent\textbf{Type of Conditions.} On top of the baseline network, we first investigate a single-scale variant of our network, \ie, removing filter generators $G_1$ and $G_2$ from Figure \ref{fig:structure}. As a result, feature $E_1$ and $E_2$ remain unchanged and are cast back to restoration branch via skip connections. By applying different types of conditions, we observe a significant improvement on average PSNR over the baselines. For example, model with image condition (Table \ref{tab:ablation}(c)) and the one with  the PSF condition (Table \ref{tab:ablation}(d)) improve $0.72$ dB and $1.27$ dB, respectively. Besides, combining both PSF and image conditions (Table \ref{tab:ablation}(e)) brings additional improvements ($1.18$/$0.63$ dB increase on PSF/image conditions). This indicates even the simplest single-scale dynamic convolution design could benefit the feature refinement.

\noindent\textbf{Single-scale \vs Multi-scale.} By applying multi-scale dynamic filter generators to transform skip connections at all scale, our proposed DISCNet (Table \ref{tab:ablation}(f)) increase $0.5$ dB over its single-scale counterpart (Table \ref{tab:ablation}(e)). This demonstrates the effectiveness of multi-scale strategy.

\noindent\textbf{Size of Dynamic Filters.} To further investigate the best trade-offs between performance and model size, we vary the size of dynamic filters. As shown in Table \ref{tab:size}, larger size of filters can bring better performance. However, the performance become even worse by increasing size after $s=5$, while the amount of parameters significantly increases. Hence, we empirically choose $s=5$ by default.

\begin{table}[t]
  \centering
  \caption{Results over different sizes of dynamic filters.}
  \label{tab:size}
  \begin{tabular}{lcccc}
  	\hline\hline
  	Filter Size & $s=3$ & $s=5$ & $s=7$ & $s=9$ \\\hline
    PSNR & $42.16$ & $42.77$ & $42.62$ & $42.47$ \\
    SSIM & $0.9862$ & $0.9870$ & $0.9869$ & $0.9868$ \\
    LPIPS \cite{zhang2018unreasonable} & $0.0126$ & $0.0119$ & $0.0119$ & $0.0119$ \\
  	\hline
  	Params (M) & $3.18$ & $3.44$ & $3.84$ & $4.37$ \\
  	FLOPs (G) & $262.10$ & $272.59$ & $288.32$ & $309.29$ \\
  	\hline\hline
  \end{tabular}
  \vskip -0.5cm
\end{table}

%------------------------------------------------------------------------
\vspace{-0.1cm}
\subsection{Evaluation on Simulated Dataset}

\begin{figure*}[t]
\begin{center}
  \includegraphics[width=0.97\linewidth]{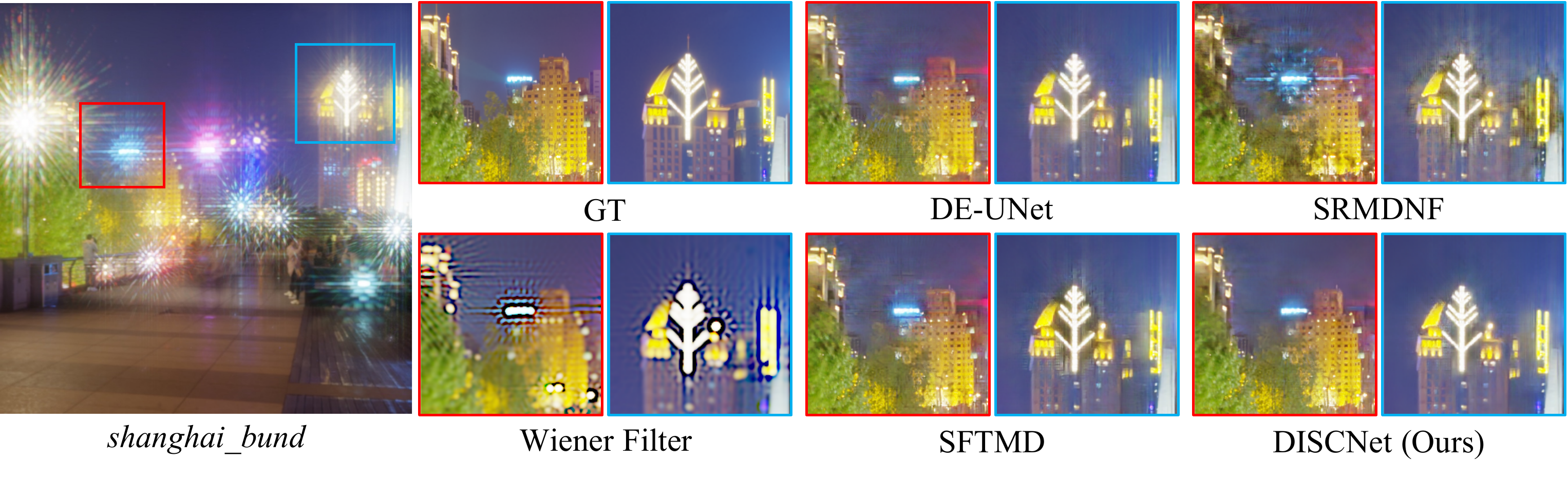}
\end{center}
\vskip -0.6cm
   \caption{\textbf{Visual comparison on a synthetic input image.} Our method restores fine details and suppresses flare effects in both highlight and dark regions and renders visually pleasing results. Refer to Supplement for more visual results. Zoom in for better view.}
\label{fig:syn_visual}
\vskip -0.3cm
\end{figure*}

\begin{figure*}[!ht]
\begin{center}
  \includegraphics[width=0.97\linewidth]{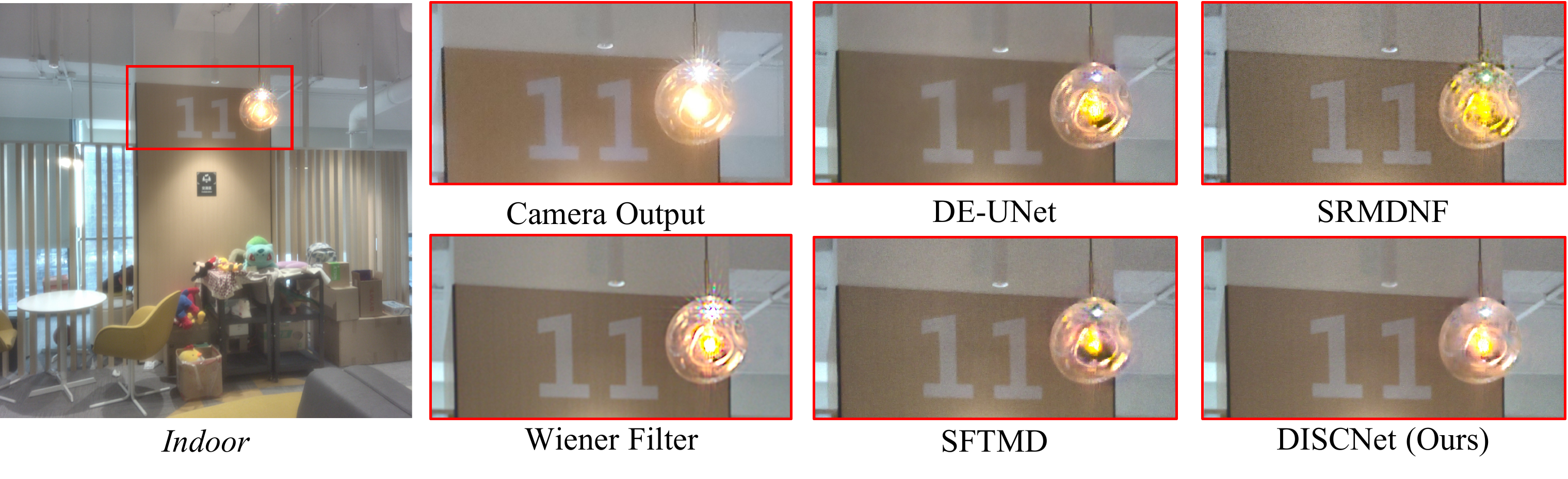}
\end{center}
    \vskip -0.6cm
   \caption{\textbf{Visual comparison on a real input image.} Our method achieves the best perceptual quality while other approaches leave noticeable artifacts and suffer from strong noise. Refer to Supplement for more visual results. Zoom in for better view.}
\label{fig:real_visual}
\vspace{-0.3cm}
\end{figure*}

To demonstrate the efficiency of DISCNet, we conduct experiments to evaluate the performance on simulated dataset. Since UDC image restoration is a newly-defined problem, we carefully select and modify four representative and state-of-the-art non-blind image restoration algorithms as baselines:
\textbf{Wiener Filter} \cite{orieux2010bayesian} is a classical deconvolution algorithm for linear convolution formation. Hence, we apply Wiener deconvolution to the degraded images with measured PSF $k$ for each channel independently in the linear domain. Note that the restored images are still evaluated and displayed in tone-mapped domain.
\textbf{SRMDNF} \cite{zhang2018learning} is a noise-free version of SRMD, which integrates non-blind degradation information to handle multiple degradations in a super-resolution network. The network contains $12$ convolution layers, each of which produces $128$ feature maps. By conventions of network designed for low-level tasks \cite{wang2018esrgan,lim2017enhanced}, we remove BN layers to stabilize the training.
\textbf{SFTMD} \cite{gu2019blind}. Iterative Kernel Correction (IKC) is originally devised for image super-resolution on blind setting. In our experiments, we employ SFTMD network which also leverages the kernel information to solve the non-blind problem. We remove the pixel shuffle upsampling layer as the input and output share the same shape in UDC restoration task.
\textbf{DE-UNet} \cite{zhou2020image}. Zhou \etal presents a Double-Encoder UNet, referred to as DE-UNet in our experiments, to recover UDC degraded images. We modify the first layers of two encoders to take $3$-channel RGB images as inputs.

\vspace{-0.1cm}
\begin{table}[t]
  \centering
  \caption{\textbf{Quantitative comparison on the simulated dataset.} ``*'' indicates blind models that do not explicitly use the information of kernel. The best two results are highlighted in \red{red} and \blue{blue}.}
  \label{tab:syn_comp}
  \scalebox{0.88}{
    \begin{tabular}{l|cc|ccc}
        \hline
        \hline
        Method                          & \begin{tabular}[c]{@{}c@{}}Params\\  (M)\end{tabular} & \begin{tabular}[c]{@{}c@{}}FLOPs \\ (G)\end{tabular} & PSNR & SSIM & LPIPS \\ \hline
        WF \cite{orieux2010bayesian}    & - & - & $27.41$ & $0.8392$ & $0.3365$ \\
        SRMDNF \cite{zhang2018learning} & $1.49$ & $951$ & $34.80$ & $0.9659$ & $0.0360$ \\
        DE-UNet* \cite{zhou2020image}     & $9.02$ & $169$ & $39.81$ & $0.9795$ & $0.0206$ \\
        SFTMD \cite{gu2019blind}        & $3.85$ & $2460$ & \blue{$42.35$} & \blue{$0.9863$} & \blue{$0.0123$} \\
        DISCNet                         & $3.80$ & $364$ & \red{$43.27$} & \red{$0.9877$} & \red{$0.0108$} \\ \hline \hline
    \end{tabular}}
  \vskip -0.5cm
\end{table}

% \begin{packed_itemize}
%     \item \textbf{Wiener Filter} \cite{orieux2010bayesian} is a classical deconvolution algorithm for linear convolution formation. Hence, we apply Wiener deconvolution to the degraded images with measured PSF $k$ for each channel independently in the linear domain. Note that the restored images are still evaluated and displayed in tone-mapped domain.
% %
%     \item \textbf{SRMDNF} \cite{zhang2018learning} is a noise-free version of SRMD, which integrates non-blind degradation information to handle multiple degradations in a super-resolution network. The network contains $12$ convolution layers, each of which produces $128$ feature maps. By conventions of network designed for low-level tasks \cite{wang2018esrgan,lim2017enhanced}, we remove BN layers to stabilize the training.
% %
%     \item \textbf{SFTMD} \cite{gu2019blind}. Iterative Kernel Correction (IKC) is originally devised for image super-resolution on blind setting. In our experiments, we employ SFTMD network which also leverages the kernel information to solve the non-blind problem. We remove the pixel shuffle upsampling layer as the input and output share the same shape in UDC restoration task.
% %
%     \item \textbf{DE-UNet} \cite{zhou2020image}. Zhou \etal presents a Double-Encoder UNet, referred to as DE-UNet in our experiments, to recover UDC degraded images. We modify the first layers of two encoders to take $3$-channel RGB images as inputs.
% \end{packed_itemize}

\noindent\textbf{Quantitative Comparisons.} For all deep learning-based methods, we train them using the same training settings and data. Table \ref{tab:syn_comp} shows quantitative results on simulated dataset. The proposed algorithm performs favorably against other baseline methods.
%In addition, the exact PSNR values of different methods on different dataset induced by various PSF kernel are shown in Figure \ref{}.
%
We observe that the proposed DISCNet consistently outperforms all other approaches on the simulated dataset. Even with the exact PSF kernel, Wiener Filter \cite{orieux2010bayesian} only achieves low image quality far below that of deep learning-based methods. SRMDNF \cite{zhang2018learning} builds upon a plain network and uses a simple strategy to utilize the kernel information. Therefore, it cannot adapt to degraded regions caused by highlight sources and produces inferior results. Compared to SFTMD \cite{gu2019blind}, our network could achieve better performance with only $15\%$ computational cost (decline from $2459.57$ to $364.34$ GFLOPs). This suggests DISCNet is efficient and particularly fit for this task, while any other boiler-plate network (\eg, plain net, UNet) produces unsatisfactory results.

\noindent\textbf{Visual Comparisons.} Figure \ref{fig:syn_visual} compares the proposed model with existing methods on simulated dataset. As one can see, Wiener filter produces unpleasing results and suffers from serious ringing and halo artifacts. In comparison, our DISCNet generates the most perceptually pleasant results and removes diffraction artifacts derived from highlights in the unsaturated regions.
The presented visual results in Figure \ref{fig:fig_1} and Figure \ref{fig:syn_visual} and additional results in the Supplemental Material validate the performance of the proposed DISCNet for various scene types, \eg, night-time urban scenes and indoor settings with strong light sources.

%------------------------------------------------------------------------
\vspace{-0.1cm}
\subsection{Evaluation on Real Dataset}
\vspace{-0.1cm}
Apart from the evaluation on synthetic dataset, this section explores reconstruction performance on real dataset. Since the ground-truth images are inaccessible, we provide the qualitative comparisons as shown in Figure \ref{fig:real_visual}. We also include the camera output of ZTE phone for comparisons. As the real data is captured without ISP, we adopt simple post-processing to all outputs except camera output for better visualization. Our network achieves the best perceptual quality while other approaches leave noticeable artifacts and suffer from strong noise or flare. Post-processing and more visual results can be found in Supplement.

%-------------------------------------------------------------------------
%----------- Discussion --------------------------------------------------
%-------------------------------------------------------------------------
\vspace{-0.2cm}
\section{Discussion}
\noindent\textbf{Limitations.}
%Under the non-blind setting, our method assumes the awareness of PSF, which is impractical in real scenarios. Kernel mismatch could severely deteriorate the performance of the network. In addition, while achieving rather satisfactory results on simulated data, DISCNet still struggles with domain discrepancy in real data and in some cases generates over-corrected results. Please refer to the Supplemental Material for further discussion and failure cases. By no means we solve the restoration problem of diffraction artifacts. This work, however, still outlines some possible avenues for future exploration.
%
Our work is only the first step towards removing diffraction image artifacts in UDC systems. Other complexities, \eg, spatially-varying PSF, noise in low light, and defocus, require more study. The proposed DISCNet sometimes will fail due to the domain gap between simulated and real data, \eg, camera noise, motion blur, variations in scenes. Our method currently is also too heavy-weight. See Supplement for further discussion and failure cases.

\noindent\textbf{Conclusion.} In this paper, we define a physics-based image formation model and measure the real-world PSF of the UDC system, and provide a model-based data synthesis pipeline to generate realistically degraded images. Then, we propose a new domain knowledge-enabled Dynamic Skip Connection Network (DISCNet) to restore the UDC images.
We offer a foundation for further exploration in UDC image restoration. Our perspective on UDC has potential to inspire more diffraction-limited image restoration work.

%-------------------------------------------------------------------------
%----------- Ack --------------------------------------------------
%-------------------------------------------------------------------------
\noindent\textbf{Acknowledgements.}
\footnotesize{Thanks Joshua Rego and Guiqi Xiao for assistance on data collection. This research was conducted in collaboration with SenseTime. This work is supported by A*STAR through the Industry Alignment Fund - Industry Collaboration Projects Grant.
}

% \clearpage
{\small
\bibliographystyle{ieee_fullname}
\bibliography{egbib}
}

\end{document}

%%%%%%%%% TITLE
\title{Removing Diffraction Image Artifacts in Under-Display Camera via \\Dynamic Skip Connection Network \\Suppplementary Material}

\author{
Ruicheng Feng$^{1}$\quad
Chongyi Li$^{1}$\quad
Huaijin Chen$^{2}$\quad
Shuai Li$^{2}$\quad
Chen Change Loy$^{1}$\quad
Jinwei Gu$^{2,3}$
\\
$^{1}$S-Lab, Nanyang Technological University\quad
$^{2}$Tetras.AI\quad
$^{3}$Shanghai Al Laboratory\\
{\tt\small \{ruicheng002, chongyi.li, ccloy\}@ntu.edu.sg}
\\
{\tt\small \{huaijin.chen, shuailizju\}@gmail.com\quad
gujinwei@tetras.ai}
}

\maketitle

%%%%%%%%% Content
In this Supplementary Material, we present additional details and discussions for the image formation model and DISCNet proposed in the main body as follows:
\begin{itemize}
    \item Light Propagation Model
    \item Incomplete Degradation in LDR Scenes
    \item Comparison with Previous Dataset
    \item Training Details
    \item Limitations
    \item Visual Results on Simulated Dataset
    \item Visual Results on Real Dataset
\end{itemize}

%%%%%%%%% BODY TEXT
%-------------------------------------------------------------------------

\section{Light Propagation Model}
\label{sec:psf}

The light propagation model in the UDC system can be divided into following steps:

\noindent\textbf{Propagation between the point source and the OLED display.} The light emitted from the point light source first hit on the front plane of the OLED display, where the optical field $U_{D-}(p,q)$ can be expressed as
\begin{equation}
\label{eqn:prop1}
    U_{D-}(p,q)=\exp\left({\frac{i\pi}{\lambda z_{1}}(p^2+q^2)}\right),
\end{equation}
where $(p,q)$ is the 2D spatial coordinates, $\lambda$ is the wavelength and $z_{1}$ is the distance between the point light source and the OLED display. We assume that the point source has unit amplitude.

\noindent\textbf{Modulation by the OLED display.} The light hit on the front plane of the OLED display will be modulated by its transmission function $t(p,q)$, which is determined by the specific design of the display pattern. The optical field after modulation $U_{D+}(p,q)$ becomes
\begin{equation}
\label{eqn:prop2}
    U_{D+}(p,q)=U_{D-}(p,q)t(p,q).
\end{equation}

\noindent\textbf{Propagation between the OLED display and the lens.} The light modulated by the OLED display propagates for a distance of $d$, before hitting on the front plane of the lens, where the optical field $U_{L-}(p,q)$ can be computed using Fresnel propagation as
\begin{equation}
\label{eqn:prop3}
    U_{L-}(p,q)=U_{D+}(p,q)*\exp\left({\frac{i\pi}{\lambda d}(p^2+q^2)}\right).
\end{equation}
Here, $*$ denotes the 2-D convolution operator.

\noindent\textbf{Modulation by the lens.} The light hit on the front plane of the camera lens will be modulated by the lens transmission function, which is determined by focal length $f$ of the lens. The optical field after modulation $U_{L+}(p,q)$ becomes
\begin{equation}
\label{eqn:prop4}
    U_{L+}(p,q)=U_{L-}(p,q)\exp\left({\frac{-i\pi}{\lambda f}(p^2+q^2)}\right).
\end{equation}

\noindent\textbf{Propagation between the lens and the sensor.} The light modulated by the lens propagates for a distance of $z_{2}$, before hitting on the sensor, where the optical field $U_{S}(p,q)$ can be computed using Fresnel propagation as
\begin{equation}
\label{eqn:prop5}
    U_{S}(p,q)=U_{L+}(p,q)*\exp\left({\frac{i\pi}{\lambda z_{2}}(p^2+q^2)}\right).
\end{equation}

Finally, the PSF of the imaging system is given by
\begin{equation}
    k=|U_{S}|^{2}.
\end{equation}

\begin{figure}[t]
\centering
  \includegraphics[width=0.98\linewidth]{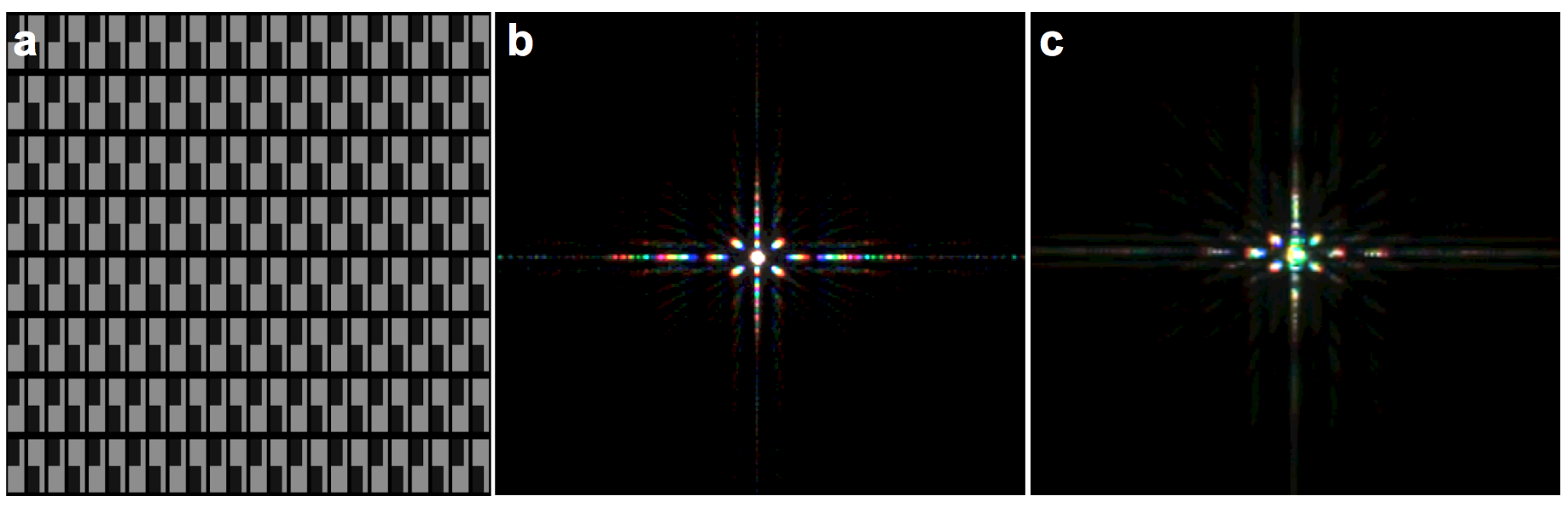}
\caption{\textbf{Comparison of simulated and real-measured PSF.} (a) The pixel layout of a commercial OLED display. Here, different gray-scale values represent different light transmittance of the display. (b) Simulated PSF. (c) Real-measured PSF. The PSFs are brightened to visualize the structured sidelobe patterns.}
\label{fig:simu}
% \vskip -0.5cm
\end{figure}

With the above equations, we can theoretically simulate the PSF of a UDC system, given the exact pixel layout of a display. Due to proprietary reasons, we do not have access to the detailed pixel structures of the particular UDC device (ZTE Axon 20) that we used in the main paper. To validate the above light propagation model, we place another commercial OLED display with known pixel layout in front of a normal camera to construct a UDC system, and use it to measure the PSF. In Figure \ref{fig:simu}, we found that although the simulated and real-measured PSF share a similar shape, they slightly differ in color and contrast due to model approximations and manufacturing imperfections.

%-------------------------------------------------------------------------
\section{Incomplete Degradation in LDR Scenes}
As described in Section 3.2 of the main body, images captured by UDC systems in real HDR scenes will exhibit structured flares near strong light sources. Since the PSF of UDC has a strong response at the center but vastly lower energy at long-tail sidelobes, only when convolved with sufficiently high-intensity scenes, these spike-shaped sidelobes can be amplified to be visible in the degraded image.

\begin{figure}[t]
\centering
  \includegraphics[width=0.98\linewidth]{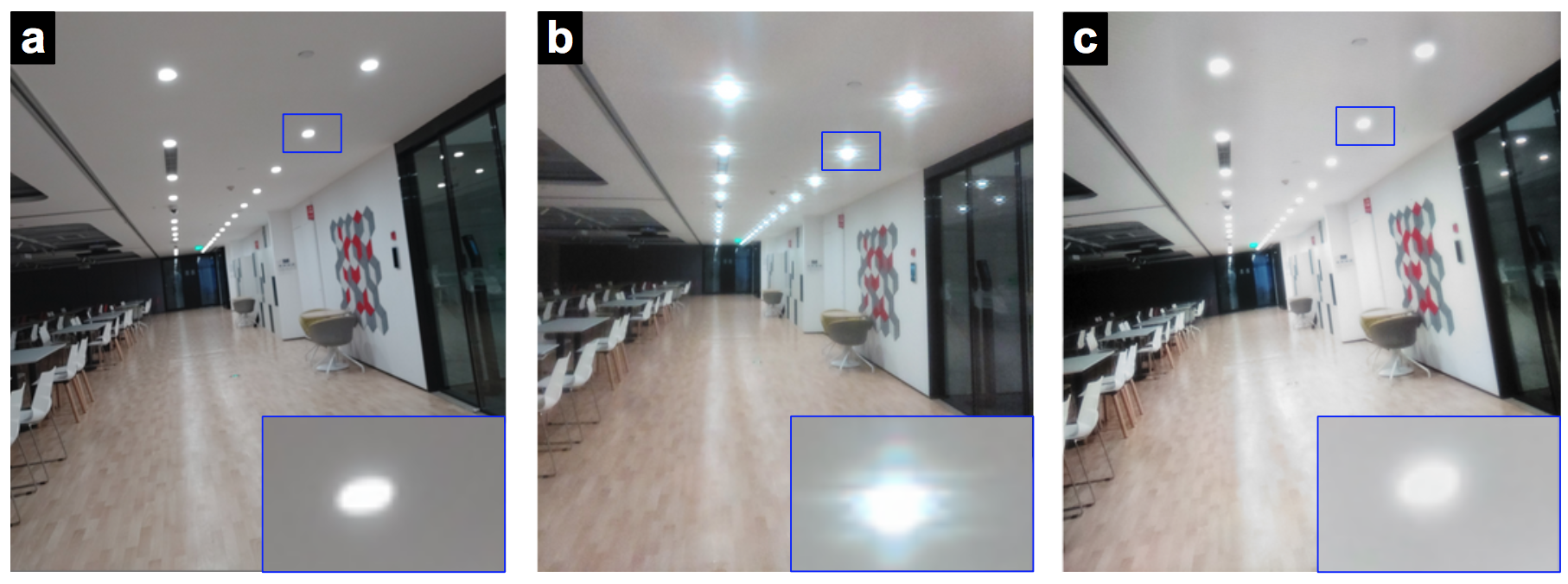}
\caption{\textbf{Comparison of UDC images of real HDR scene and monitor-generated LDR scene.} (a) Real HDR scene captured by a normal camera. (b) Real HDR scene captured by the UDC device. (c) Monitor-generated LDR scene, \ie display the image (a) on a LCD monitor, captured by the UDC device.}
\label{fig:monitor}
% \vskip -0.2cm
\end{figure}

Therefore, in an MCIS system proposed in \cite{zhou2020image}, where scenes are displayed on a LCD monitor, which commonly has limited dynamic range, the degradation of a UDC imaging system is incomplete compared to the capture in real HDR scenes. As shown in Figure \ref{fig:monitor}, if the real HDR scene is directly captured with a UDC device, we can observe flare effects near strong light sources. However, for the same scene and same imaging device, the flares are no longer visible in the acquired image if it is displayed on a LCD monitor, since the scene in this case only involves limited dynamic range.

\begin{figure}[t]
\centering
  \includegraphics[width=0.98\linewidth]{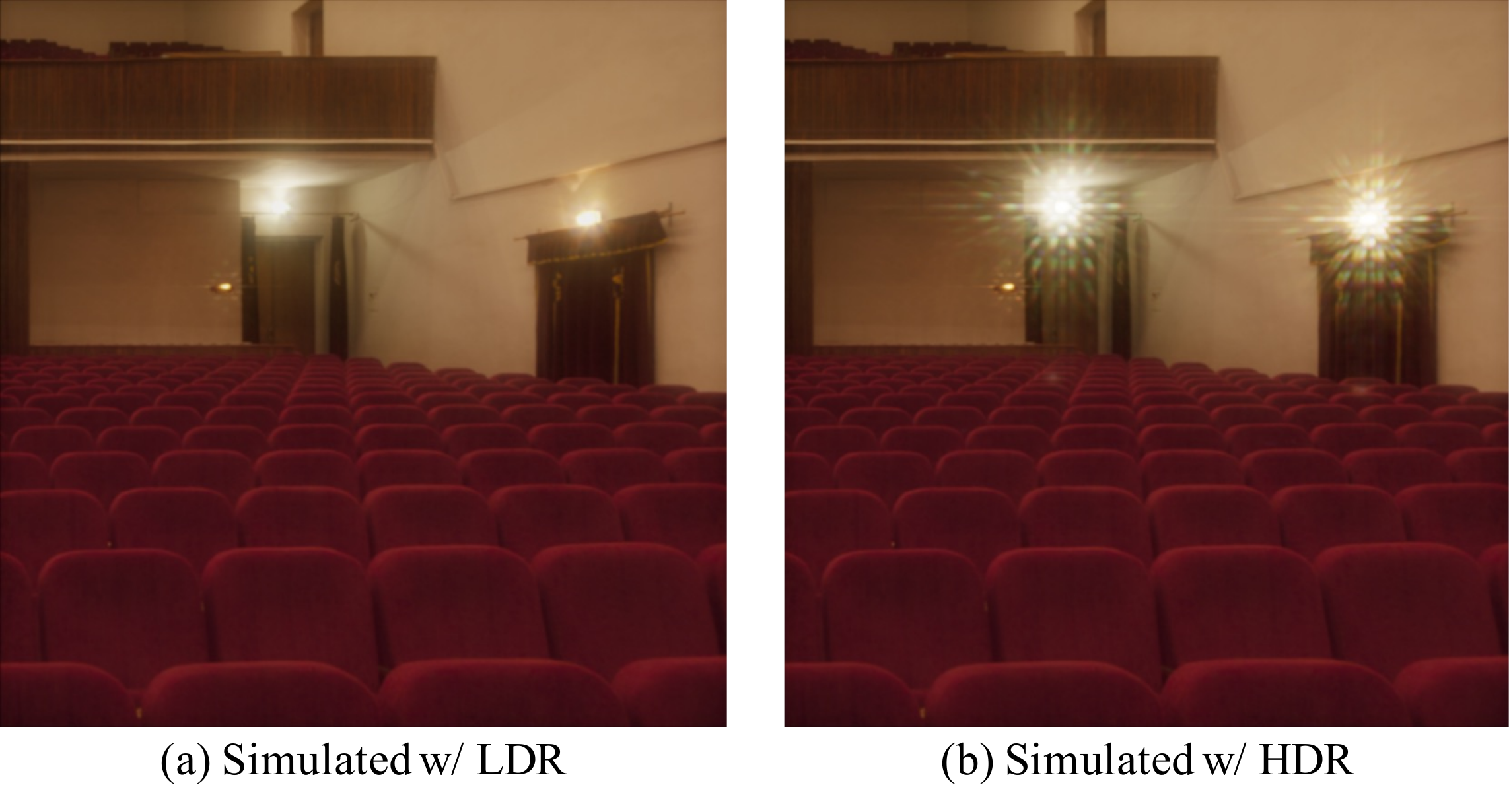}
\caption{Comparison of images simulated with LDR and HDR.}
\label{fig:simu_ldr}
% \vskip -0.2cm
\end{figure}

Apart from MCIS data, we also illustrate that HDR scenes are indispensable for our data simulation pipeline. Specifically, if we clip the scene from HDR to LDR, the flare artifacts caused by diffraction effects become invisible in the degraded images (see Figure \ref{fig:simu_ldr}). This further illustrates the importance of HDR scenes. Hence, in order to correctly model the real degradation of a typical UDC system, we involve real HDR scenes in the image formation model in main paper.

\section{Comparison with Previous Dataset}
In this section, we compare the datasets used in our work with previous one \cite{zhou2020image} in Table \ref{tab:data_comp}. The proposed image formation model could simulate more complex and realistic degradation compared to the dataset in \cite{zhou2020image}.

\begin{table*}[t]
\caption{Comparison of different datasets.}
\label{tab:data_comp}
\centering
\begin{tabular}{l|c|c|c}
\hline\hline
Dataset & Zhou \etal \cite{zhou2020image} & Simulated data (Ours) & Real data (Ours) \\\hline
Scene & Displayed on a monitor & HDR images & Real scenes \\
Dynamic Range & Low & High & High \\
Data Format & $16$-bit RAW & $32$-bit RGB & $14$-bit RGB \\
Major Degradation & Low-light, Color shift & Flare, Haze, Blur, Saturation & Flare, Haze, Blur, Saturation, Veiling glare \\
UDC System & Lab prototype & - & Commodity UDC production \\
\hline\hline
\end{tabular}
\end{table*}

%-------------------------------------------------------------------------
\section{Training Details}
\noindent\textbf{Data Simulation for Training.}
We use the image formation model in main paper to simulate degraded images exhibiting diffraction artifacts. In particular, we set $x_{max}=500$ for the clipping operation $C(\cdot)$.

For the tone mapping function, we apply a simple rule \cite{reinhard2002photographic}, given by
\begin{equation}
    \phi(x)=\frac{x}{x+\alpha},
\end{equation}
where $\alpha$ controls the scale of high luminances. The hyperparameter is set to $0.25$ in our case.
This formulation mainly compress the high intensities, scaled by approximately $1/x$, and is guaranteed to bring all intensities within displayable range. Many scenes are predominated by a normal dynamic range, but have a few high luminance regions nearby highlights, \eg, street lamp, sunlight. Besides, the sidelobes of the PSF have far lower energy compared to the main peak, leading to relatively low-intensity flare and haze effects in the degraded images. Therefore, this formulation can compress saturated highlights while preserving details in lowlight regions, providing a better display of diffraction artifacts.
For simplicity, we mainly focus on analyzing the diffraction effects of UDC and set $n=0$, providing a noise-free version of the simulated data.

For testing on simulated datasets, we build a test kernel set for quantitative evaluations of different methods. It consists $9$ selected rotation variations that are performed on the PSF, \ie, $\{-12,9,6,3,0,3,6,9,12\}$. The PSFs first rotate by an angle selected from the above set, and then are convolved with the ground-truth images using image formation model in main paper to generate the corresponding degrade images. In total each ground-truth image has $9$ degraded counterparts, yielding $9$ testing sets.
Note that only the PSF at the center of the sensor is measured, and the rest in the kernel are generated by applying rotation transformations to the center one.
Although it only considers simple variations (rotation) on the PSF, it can still be used to evaluate the performance of non-blind image restoration approaches.

\noindent\textbf{Loss Function.} To train the proposed model, we adopt two widely-used losses of image restoration tasks. We originally experimented with $\mathcal{L}_1$ loss between the reconstructed and ground-truth images in tone-mapped domain. To encourage more realistic results, we further apply the perceptual loss in \cite{johnson2016perceptual}, which is defined using the pre-trained VGG-19 network \cite{simonyan2014very} and given by
\begin{equation}
    \mathcal{L}_{VGG} = ||\Phi_l(\Tilde{x})-\Phi_l(\hat{x})||^2_2,
\end{equation}
where $\Phi_l$ is the feature maps extracted from the $l$-th layer of the pre-trained VGG network, and $\Tilde{x}$ is the reconstructed image of our network. In particular, we use the ``conv5-4'' layer as \cite{wang2018esrgan}. The total loss for training is formulated by
\begin{equation}
    \mathcal{L}_{total} = \mathcal{L}_{1} + \lambda \mathcal{L}_{VGG},
\end{equation}
where the weight $\lambda$ is set to $0.01$ for balancing the scale of different losses in our experiments.

%-------------------------------------------------------------------------
\begin{figure}[t]
\centering
  \includegraphics[width=0.98\linewidth]{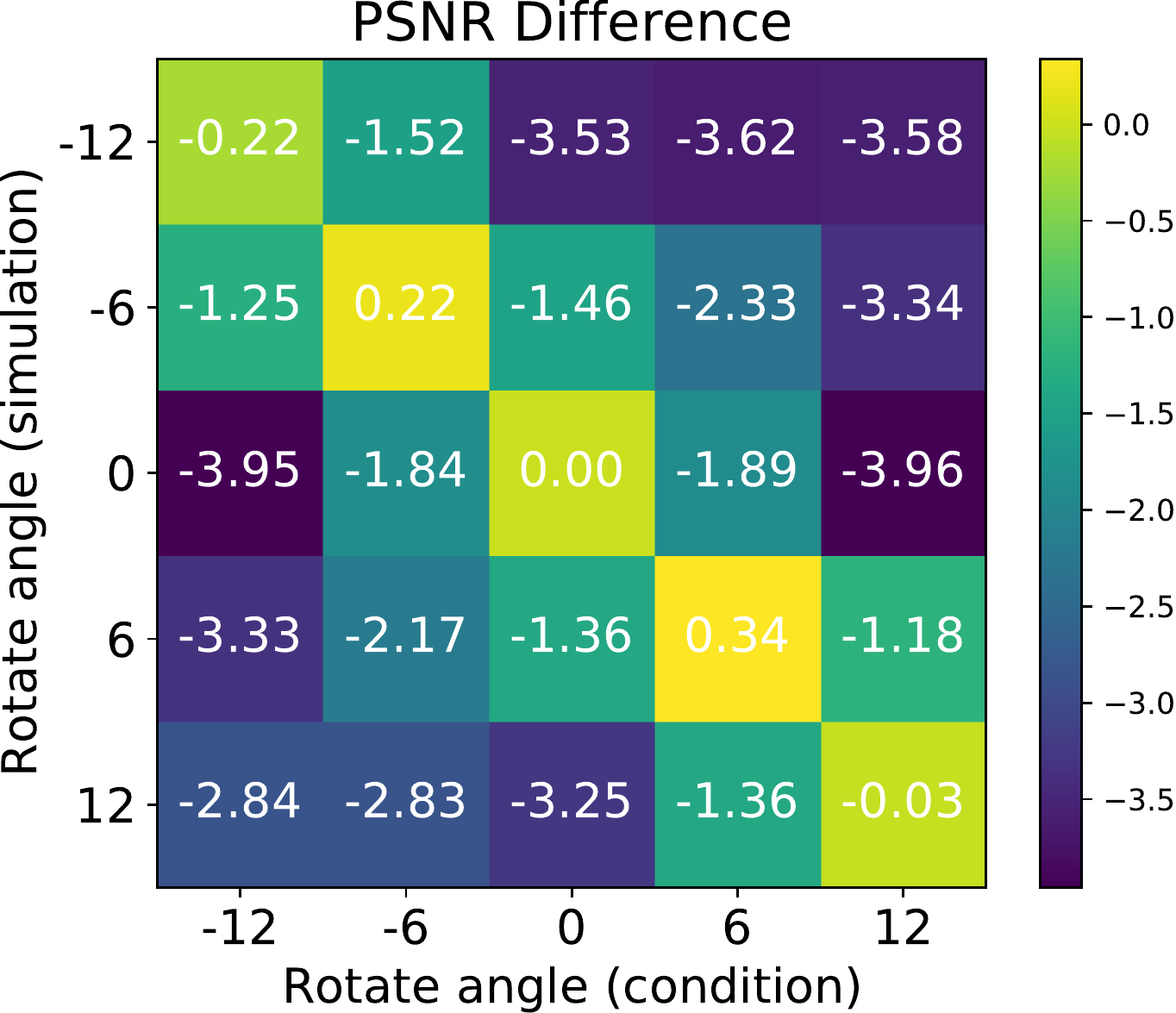}
\caption{\textbf{Recovery sensitivity to the kernel mismatch.} Simulation angles indicate rotations of PSF used in simulation, while condition angles represent the ones used as conditions to DISCNet.}
\label{fig:mismatch}
% \vskip -0.2cm
\end{figure}

\begin{figure}[t]
\centering
  \includegraphics[width=0.98\linewidth]{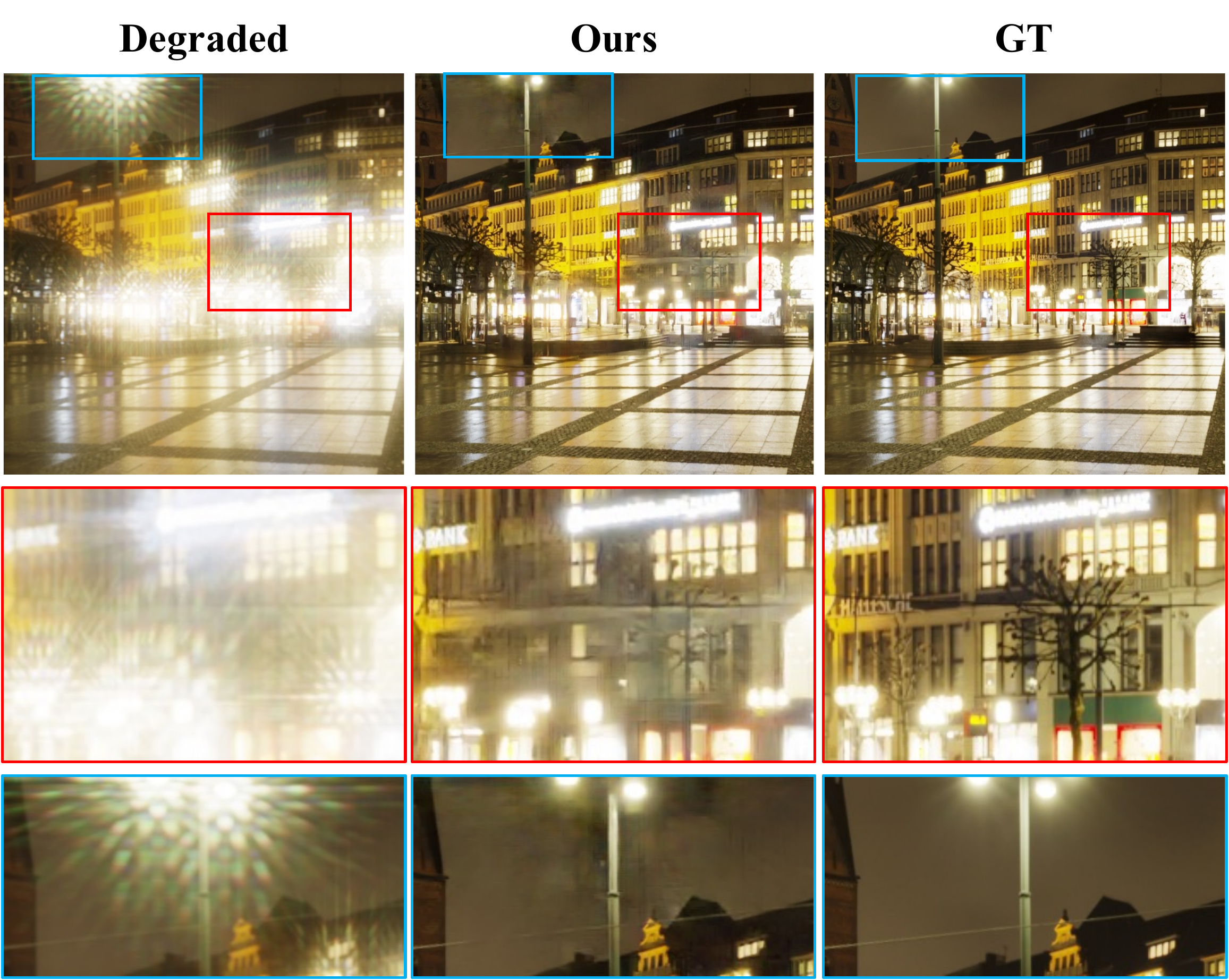}
\caption{Failure cases around strong light sources.}
\label{fig:failure_case}
% \vskip -0.2cm
\end{figure}

\section{Limitations}
\noindent\textbf{Kernel Mismatch.}
Under the non-blind setting, our method assumes the awareness of the kernel PSF. In real scenarios, however, an estimation of the PSF can be easily affected by noises and artifacts, which results in a kernel mismatch and severely deteriorates the performance of the network. Figure \ref{fig:mismatch} shows the sensitivity of the PSNR performance to the kernel mismatch. In the upper-right and lower-left regions, where the kernels used for simulation and condition differ the most, we can observe huge gaps (over 3 dB) on the PSNR performance. In contrast, the results on the diagonal, where the kernels match precisely, show the best performance in their corresponding rows or columns.

\noindent\textbf{Large and Strong Highlights.}
While achieving rather satisfactory results on small area of light sources, DISCNet still struggles when highlight regions are large and intensities are extremely strong, leading to over-corrected results. We also conduct experiments on scenes with larger and strong highlights to illustrate the limitations of our method, which is shown in Figure \ref{fig:failure_case}. The degraded image contains extremely strong light sources which causes diffraction artifacts in large neighbouring low-intensity regions. Nevertheless, our method is still able to suppress flare and haze effects and recover lost details in most regions, even when there exists limited information in these regions. The middle and bottom rows illustrate a failure mode consisting of very strong highlights that affect a large unsaturated region. Our method over-corrects the flares and leaves artifacts around the street lamps. This requires further exploration on extreme cases with large and strong highlights.

%-------------------------------------------------------------------------
\section{Visual Results on Simulated Dataset}
\begin{figure*}[t]
\begin{center}
  \includegraphics[width=0.99\linewidth]{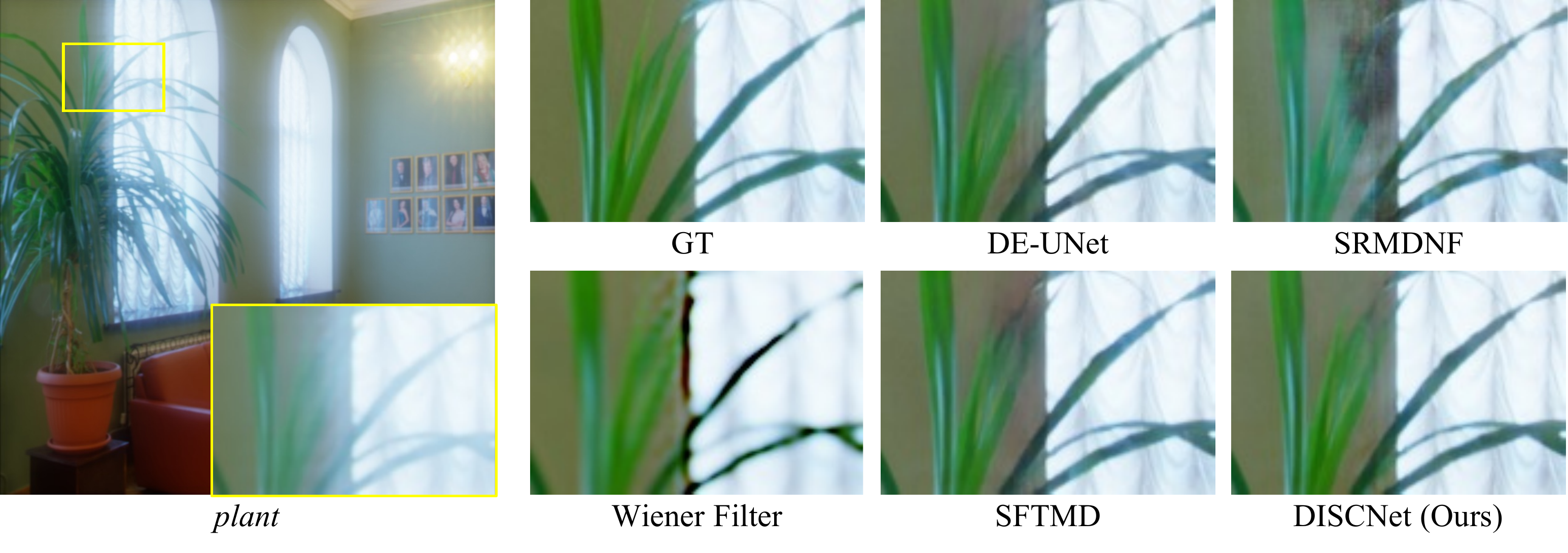}
    \vskip 0.2cm
  \includegraphics[width=0.99\linewidth]{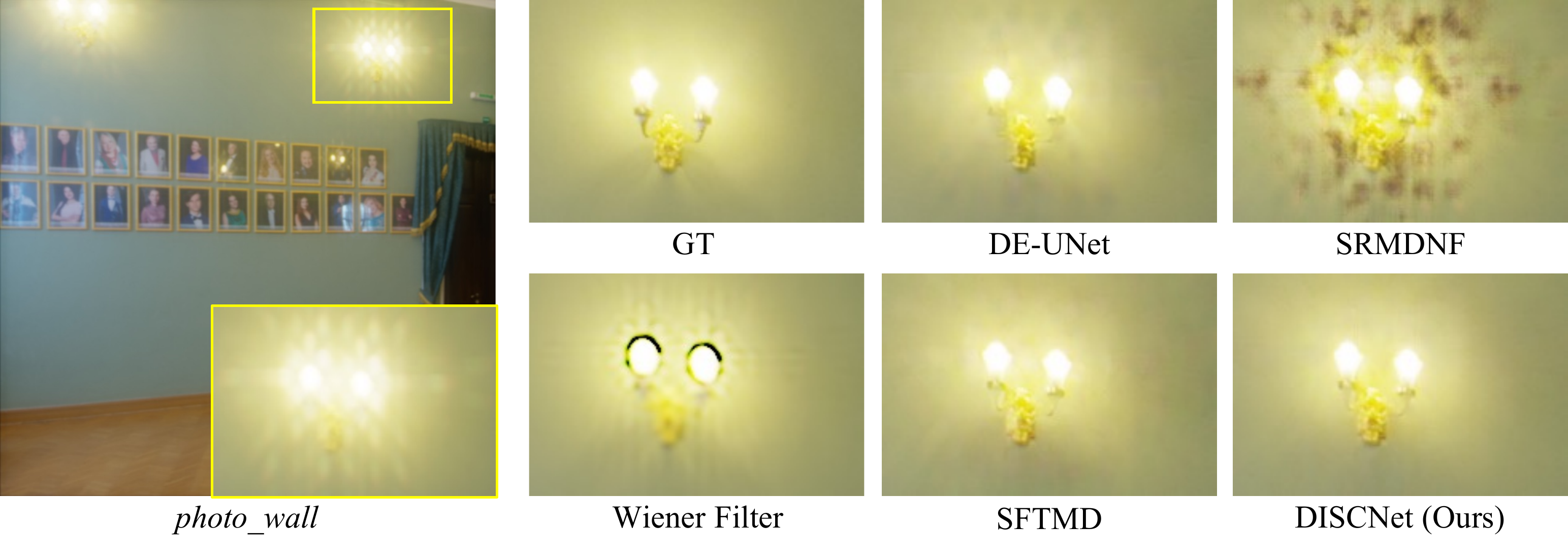}
    \vskip 0.2cm
  \includegraphics[width=0.99\linewidth]{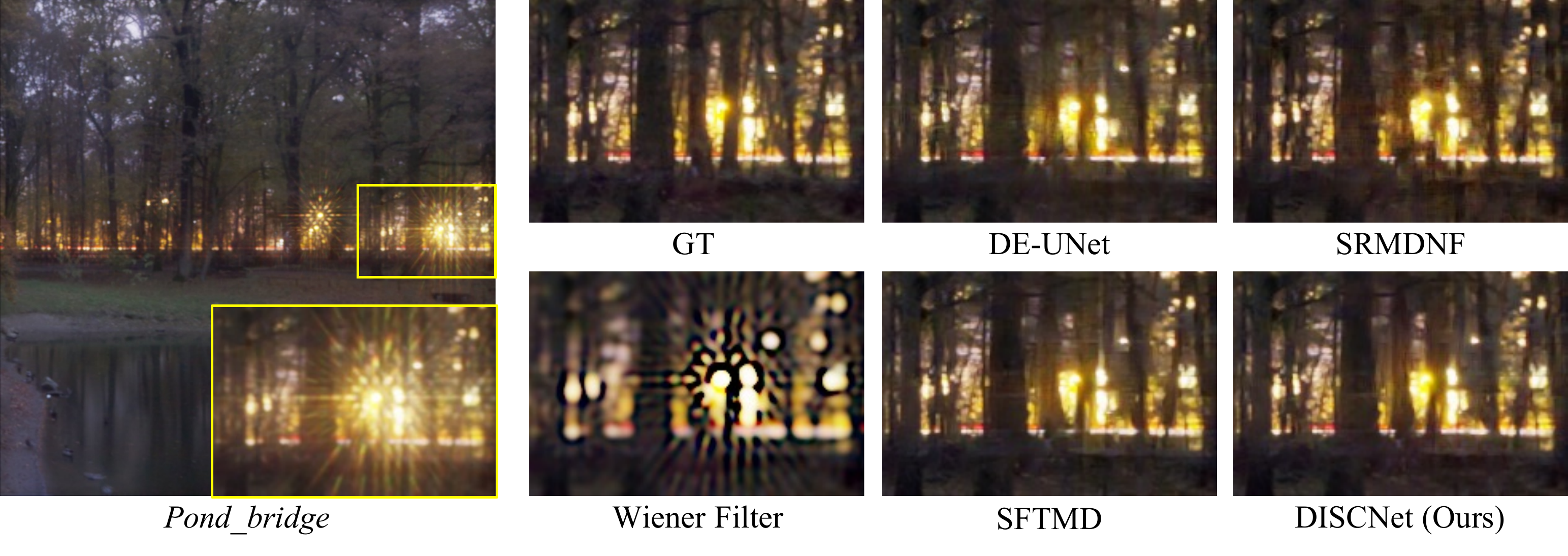}
\end{center}
\vskip -0.2cm
  \caption{Visual comparison on simulated input images. (\textbf{Zoom in for better view.})}
\label{fig:syn_visual1}
\end{figure*}

\begin{figure*}[t]
\begin{center}
  \includegraphics[width=0.99\linewidth]{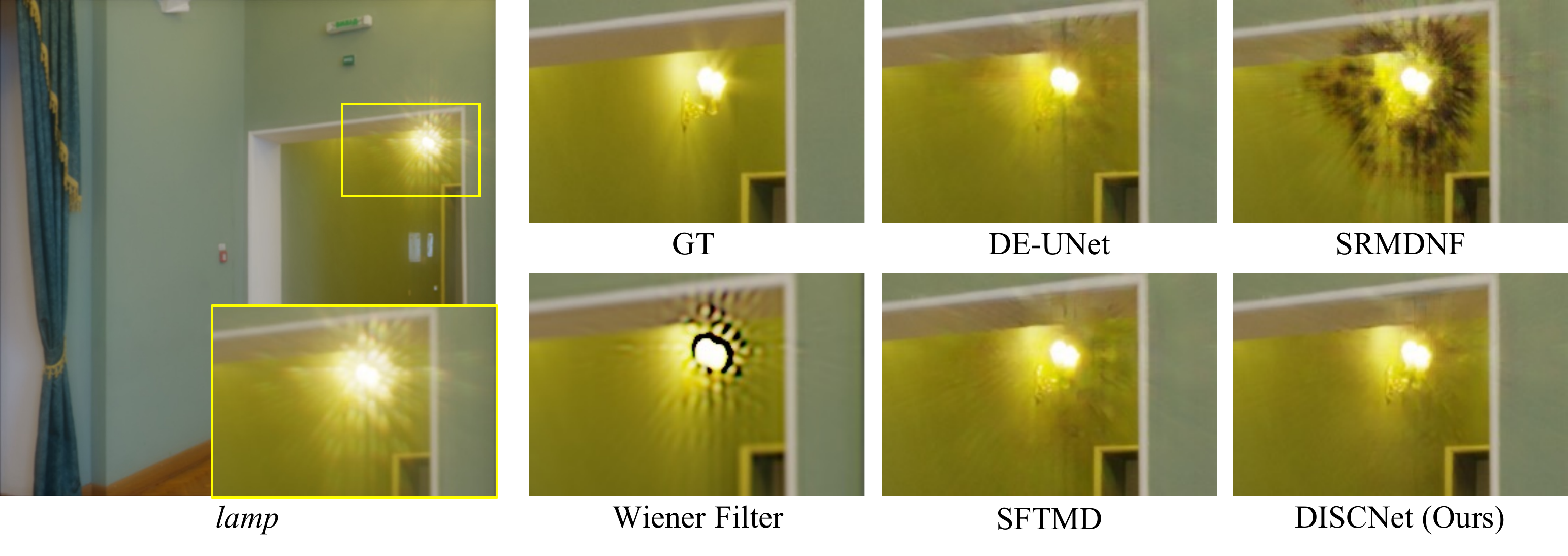}
    \vskip 0.2cm
  \includegraphics[width=0.99\linewidth]{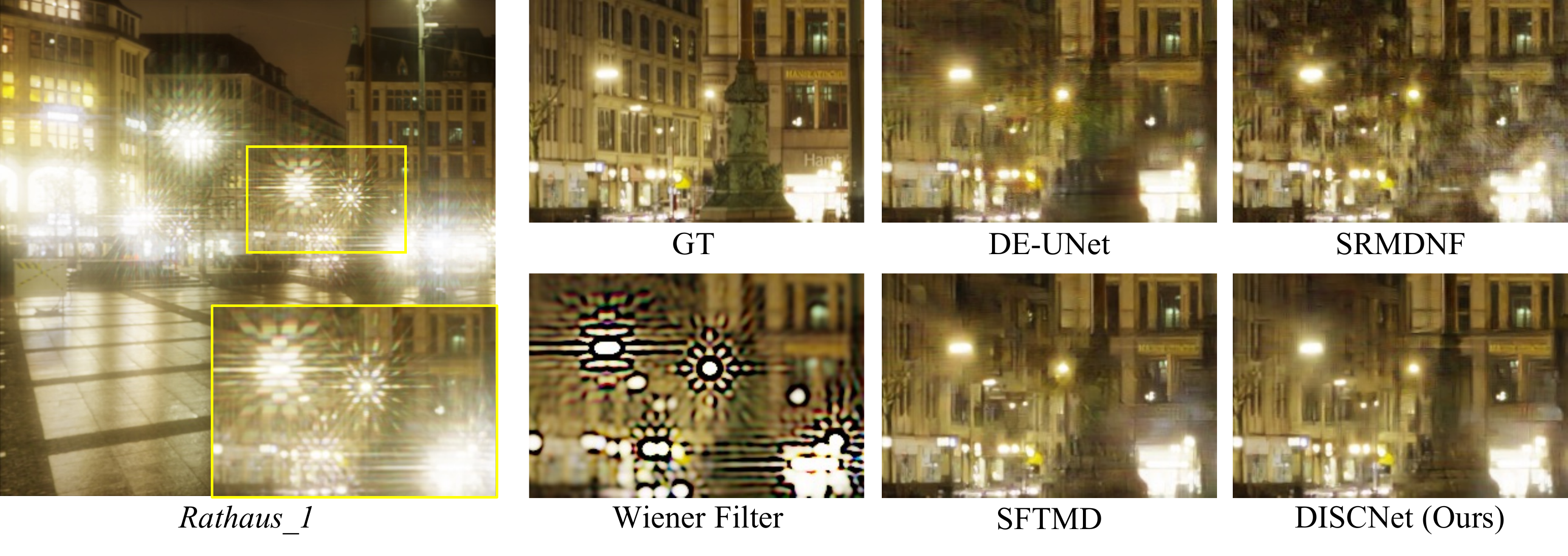}
    \vskip 0.2cm
  \includegraphics[width=0.99\linewidth]{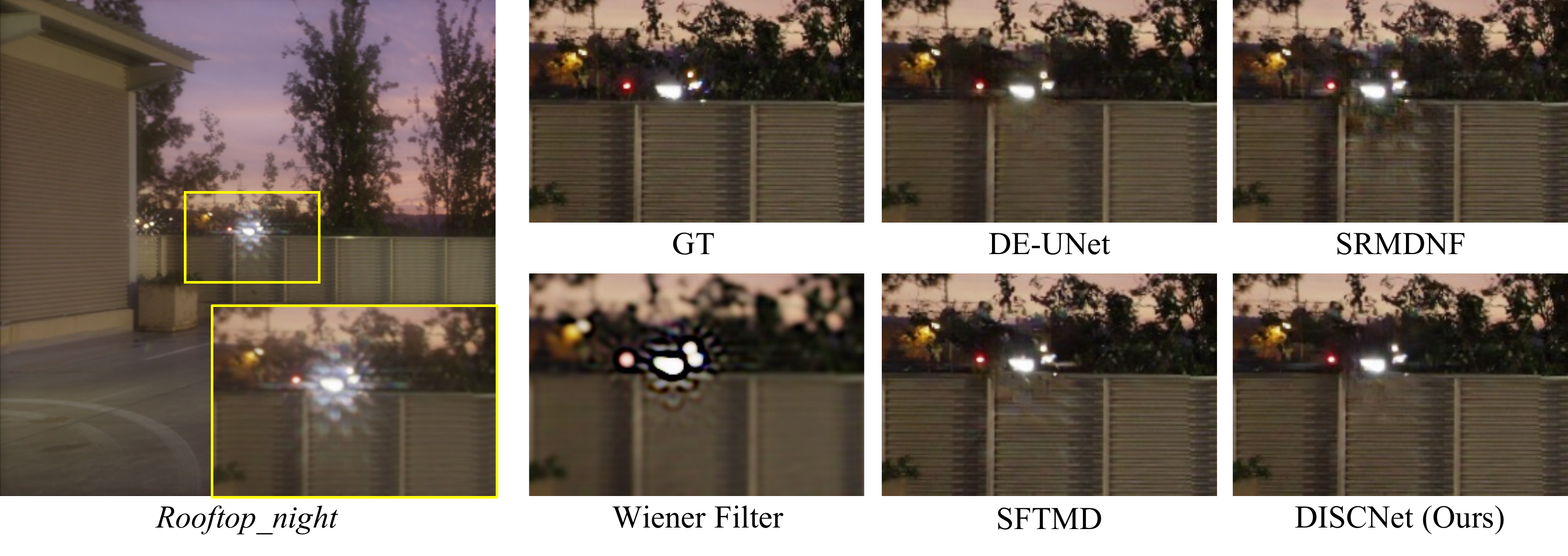}
\end{center}
\vskip -0.2cm
  \caption{Visual comparison on simulated input images. (\textbf{Zoom in for better view.})}
\label{fig:syn_visual2}
\end{figure*}

\begin{figure*}[t]
\begin{center}
  \includegraphics[width=0.99\linewidth]{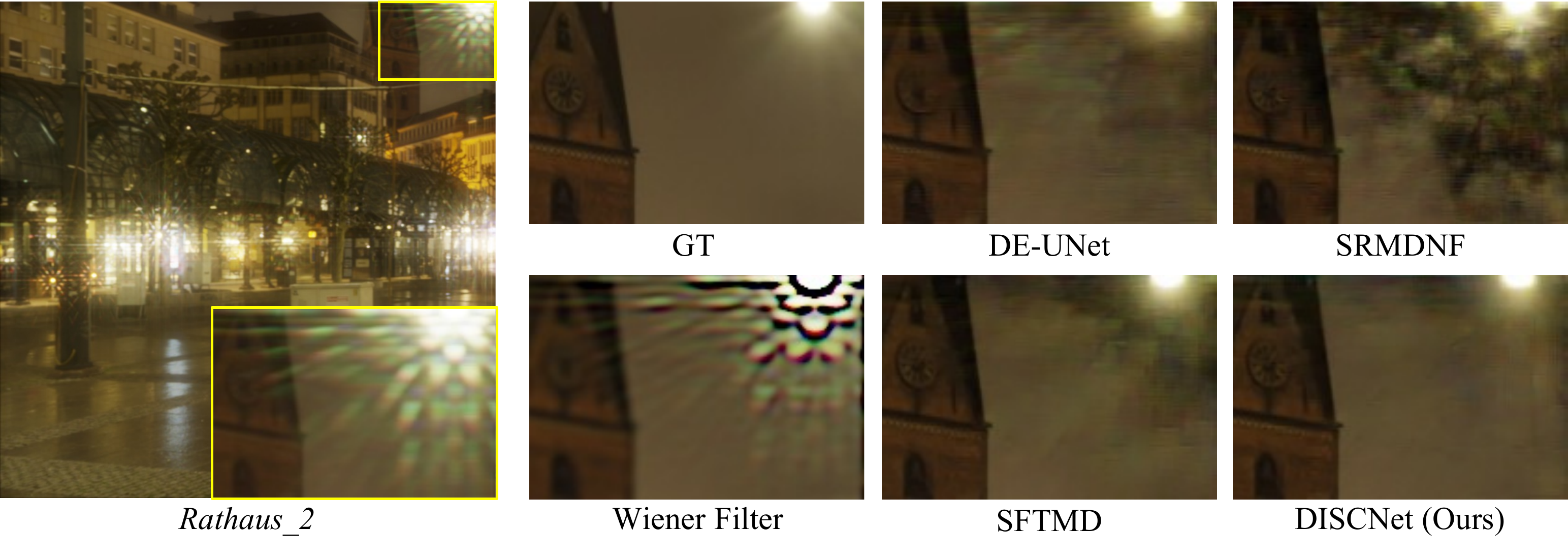}
    \vskip 0.2cm
  \includegraphics[width=0.99\linewidth]{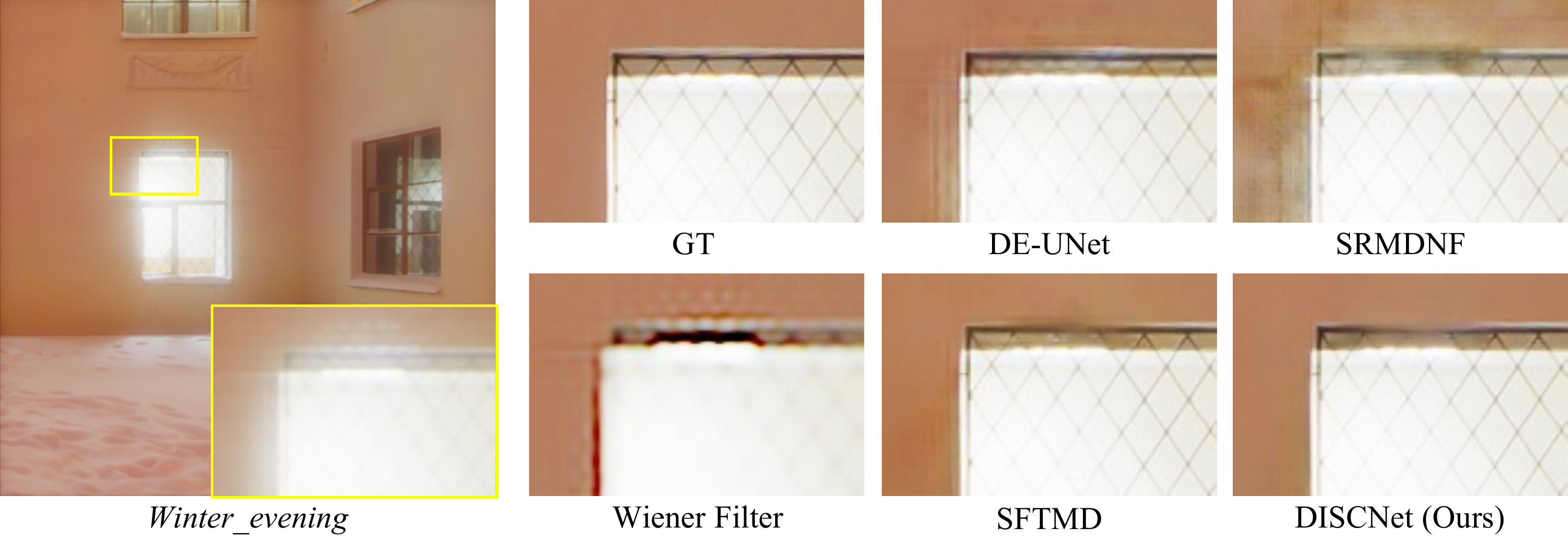}
    \vskip 0.2cm
  \includegraphics[width=0.99\linewidth]{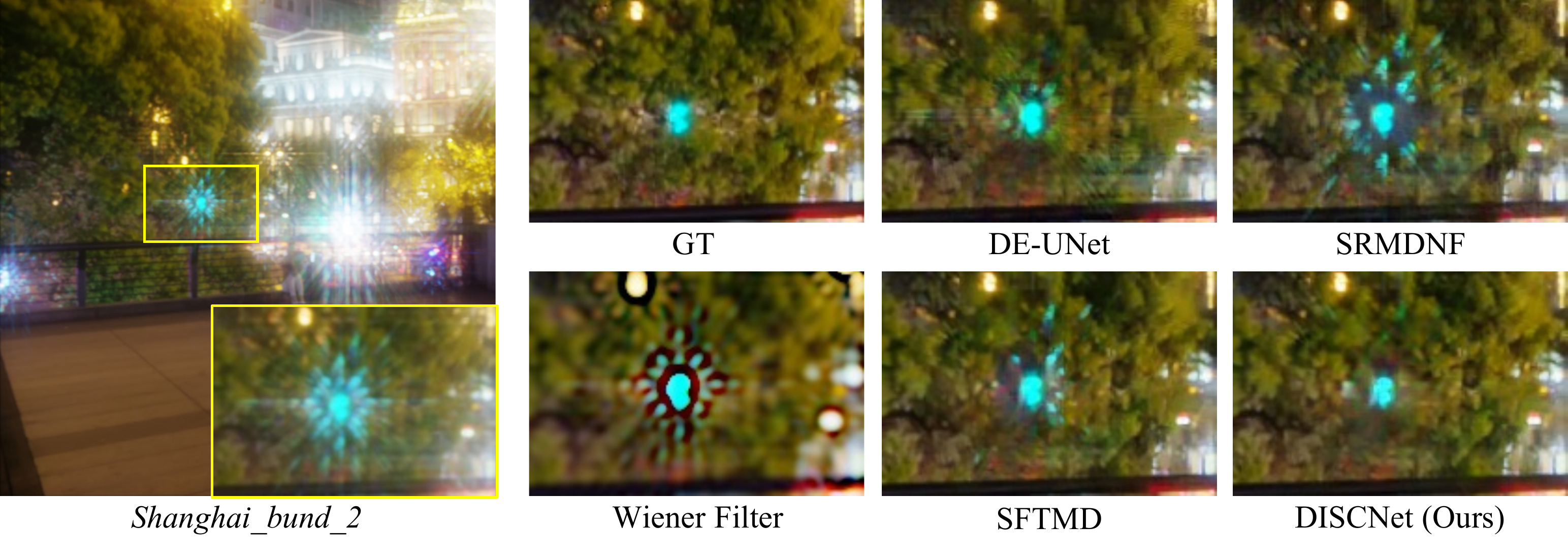}
\end{center}
\vskip -0.2cm
  \caption{Visual comparison on simulated input images. (\textbf{Zoom in for better view.})}
\label{fig:syn_visual3}
\end{figure*}

In this section, we demonstrate additional visual results on simulated data. As shown in Figure \ref{fig:syn_visual1}, Figure \ref{fig:syn_visual2}, and Figure \ref{fig:syn_visual3}, the proposed DISCNet suppresses flare and haze effects around highlights, and removes most artifacts in nearby unsaturated regions.

%-------------------------------------------------------------------------
\section{Visual Results on Real Dataset}
\noindent\textbf{Post-processing.}
Since it is beyond the scope of this paper to build a full Image Signal Processor (ISP) to output final images from raw data, we only perform a simple post-processing pipeline on the input data to adjust the color intensities and approximate the color of camera outputs, which typically exhibit perceptually better color for viewing on a display. The post-processing includes 1) color correction by color correction matrix (CCM) from the camera, 2) RGB scaling which transforms camera RGB values into camera's output RGB values, and 3) contrast enhancement using Contrast Limited Adaptive Histogram Equalization (CLAHE). Note that we also adopt the post-processing pipeline to obtain similar color in the input images for visual comparisons.

\begin{figure*}[t]
\begin{center}
  \includegraphics[width=0.99\linewidth]{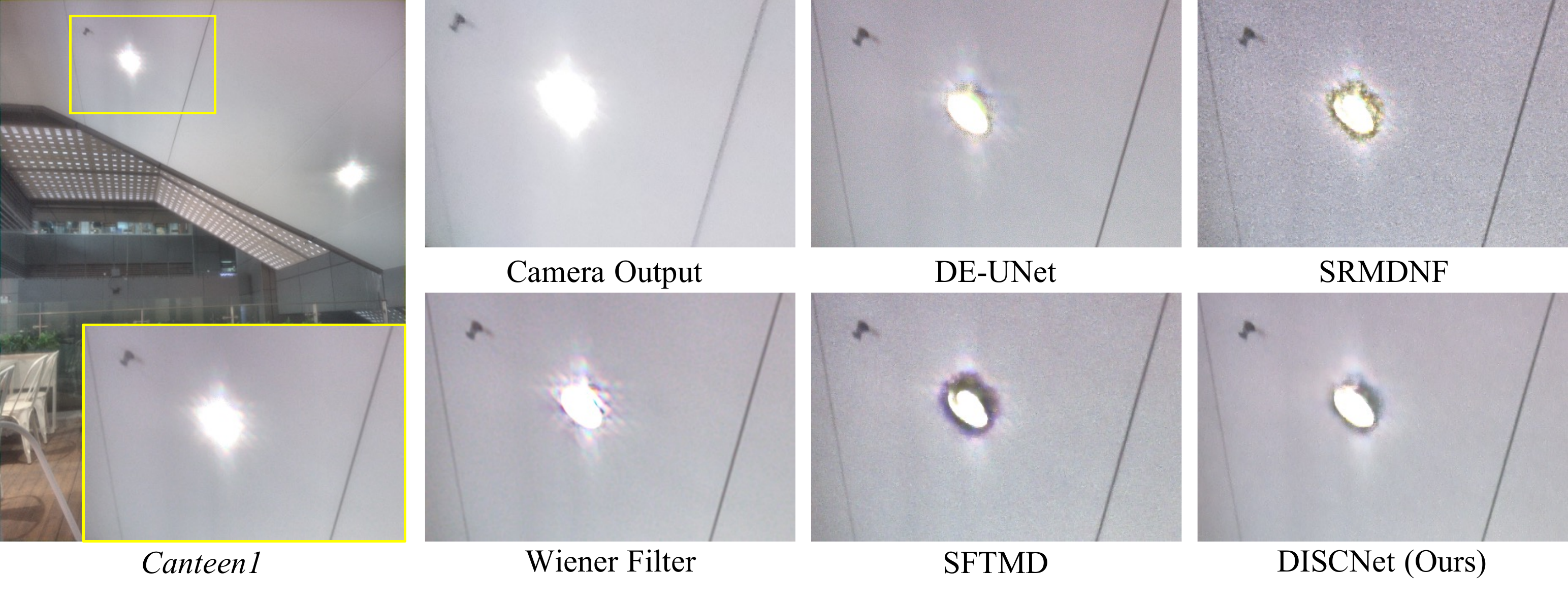}
    \vskip 0.2cm
  \includegraphics[width=0.99\linewidth]{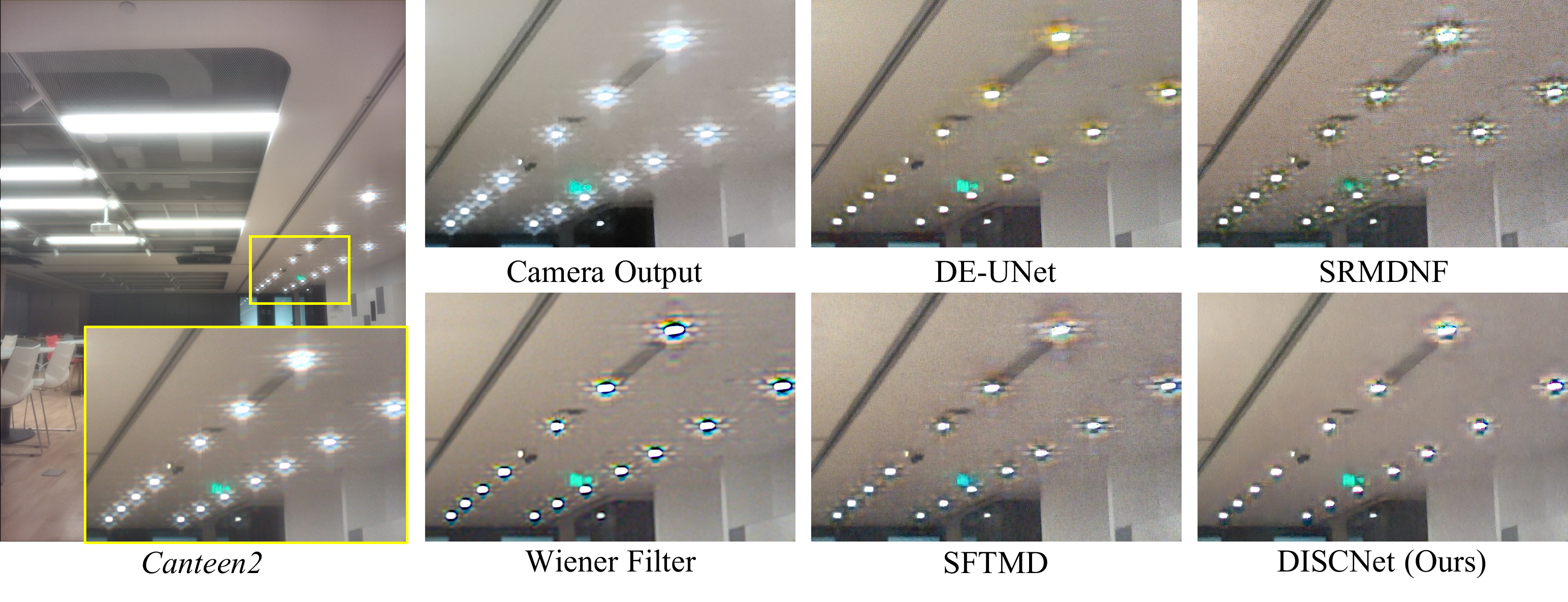}
    \vskip 0.2cm
  \includegraphics[width=0.99\linewidth]{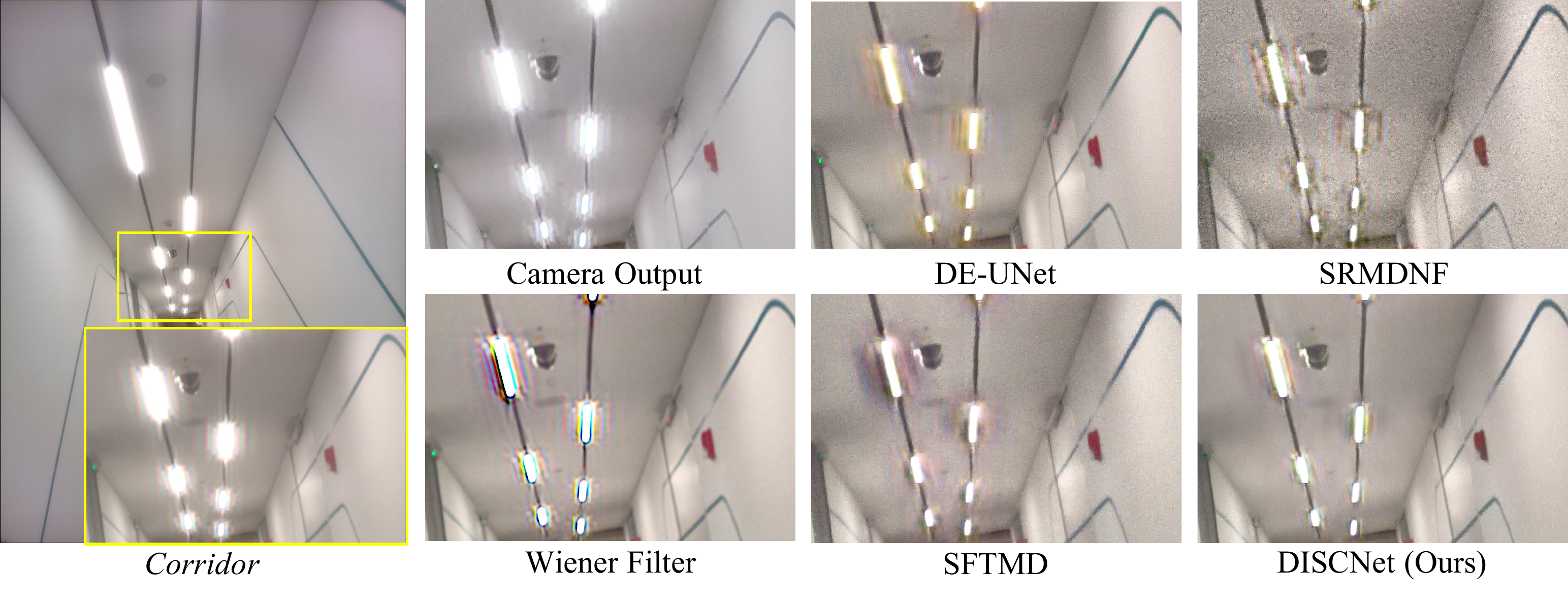}
\end{center}
\vskip -0.2cm
  \caption{Visual comparison on real input images. (\textbf{Zoom in for better view.})}
\label{fig:real_visual1}
\end{figure*}

\begin{figure*}[t]
\begin{center}
  \includegraphics[width=0.99\linewidth]{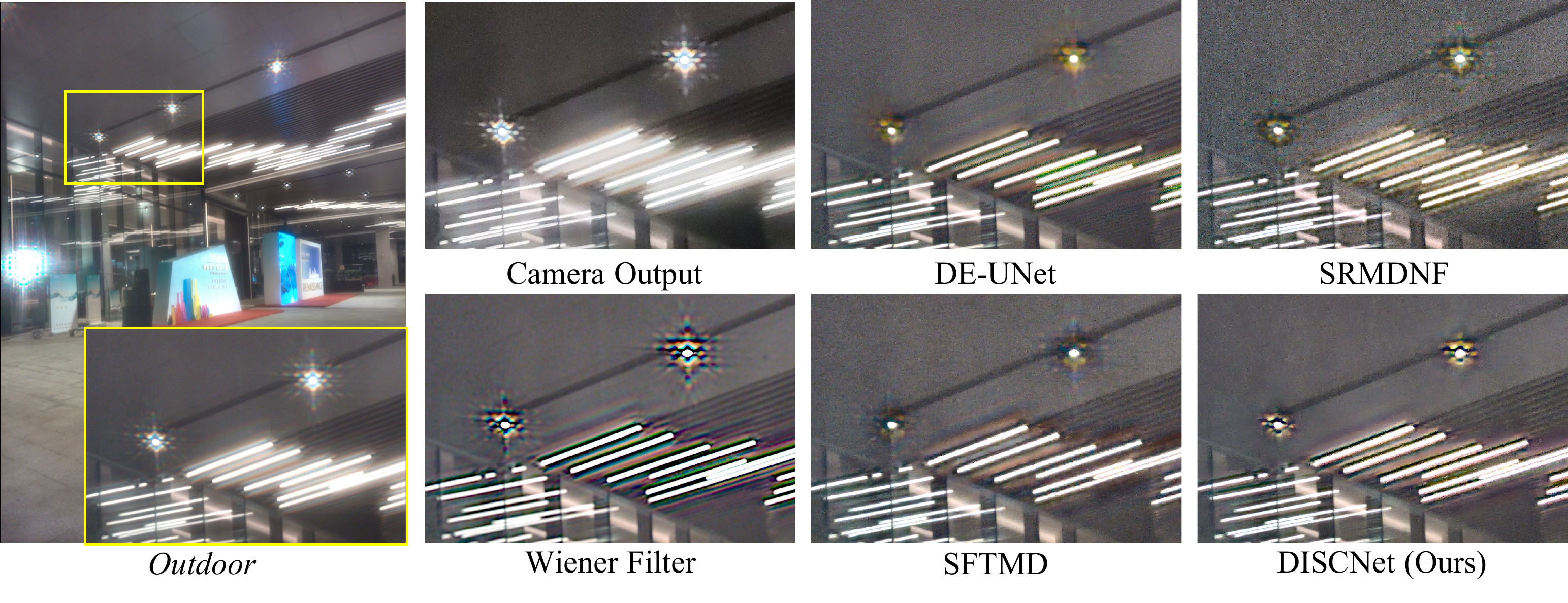}
    \vskip 0.2cm
  \includegraphics[width=0.99\linewidth]{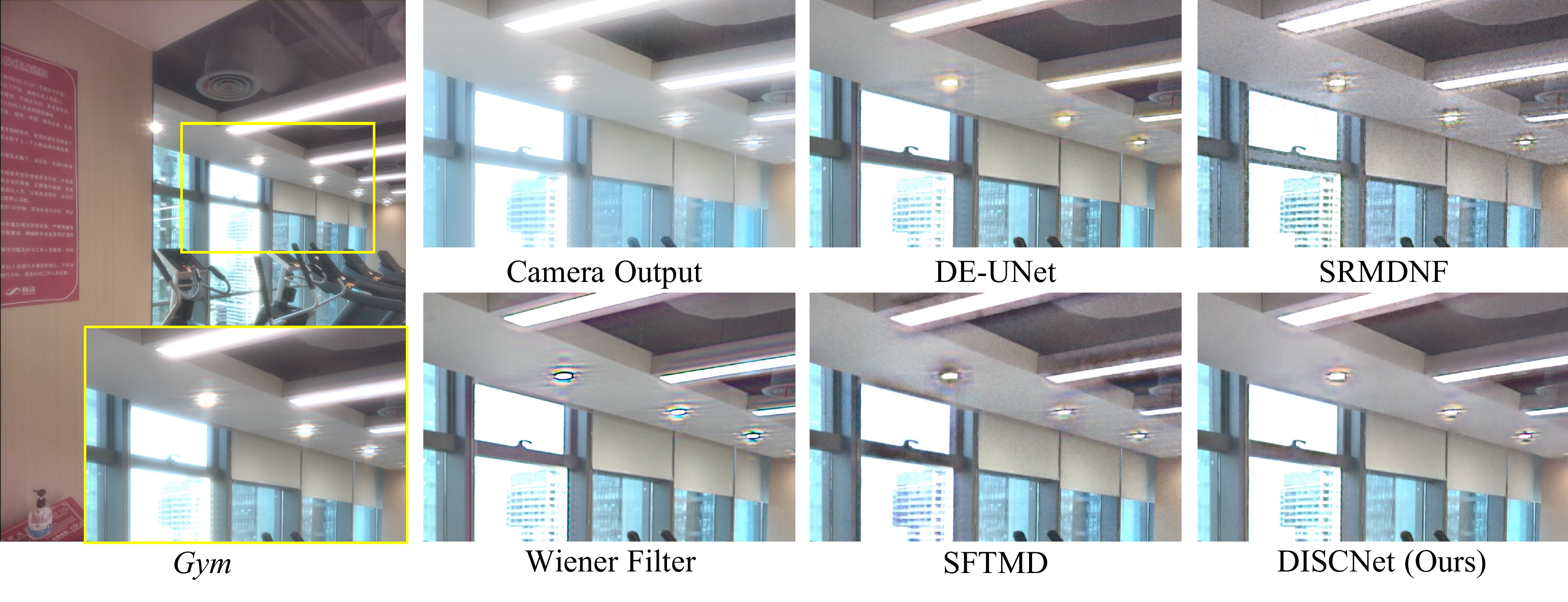}
    \vskip 0.2cm
  \includegraphics[width=0.99\linewidth]{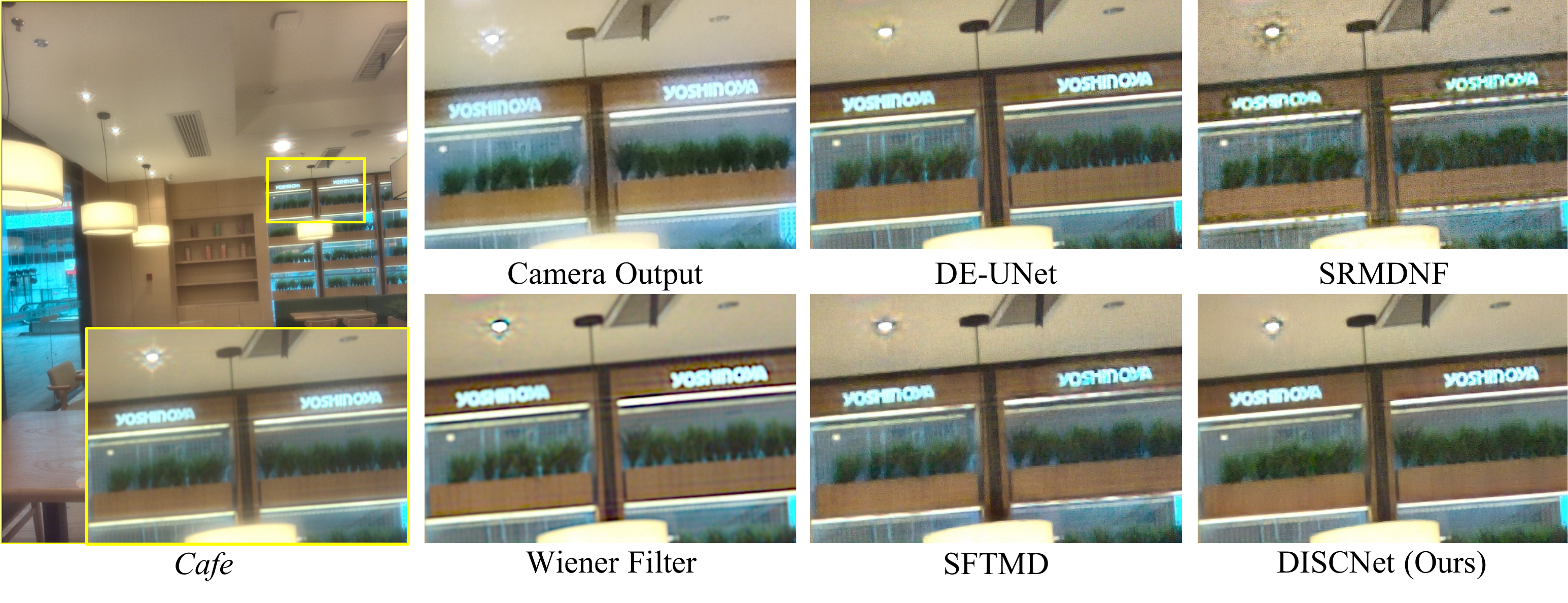}
\end{center}
\vskip -0.2cm
  \caption{Visual comparison on real input images. (\textbf{Zoom in for better view.})}
\label{fig:real_visual2}
\end{figure*}

\noindent\textbf{Visual Comparisons.}
We provide more visual comparisons with representative methods on real data in Figure \ref{fig:real_visual1} and Figure \ref{fig:real_visual2}. Our proposed network could remove diffraction image effects, while leaving least artifacts introduced by camera.

{\small
\bibliographystyle{ieee_fullname}
\bibliography{egbib}
}

%---------Simulated data------------------------------------------
\clearpage

\clearpage

\clearpage

%---------Real data------------------------------------------
\clearpage

\clearpage

% \clearpage
% {\small
% \bibliographystyle{ieee_fullname}
% \bibliography{egbib}
% }